\documentclass[sigconf]{acmart}
\renewcommand\footnotetextcopyrightpermission[1]{} 
\setcopyright{none}

\AtBeginDocument{%
  \providecommand\BibTeX{{%
    \normalfont B\kern-0.5em{\scshape i\kern-0.25em b}\kern-0.8em\TeX}}}

\usepackage{hyperref}
\usepackage{latexsym}

\usepackage{float}
\usepackage{amsfonts}
\usepackage{graphicx}
\usepackage{subcaption}

\usepackage{lipsum}

\newcommand\BibTeX{B\textsc{ib}\TeX}

\title{Generative Models are Unsupervised Predictors of Page Quality: A Colossal-Scale Study}

\author{Dara Bahri}
\authornote{Corresponding author}
\affiliation{
\institution{Google Research}
}
\email{dbahri@google.com}

\author{Yi Tay}
\affiliation{
\institution{Google Research}
}
\email{yitay@google.com}

\author{Che Zheng}
\affiliation{
\institution{Google Research}
}
\email{chezheng@google.com}

\author{Donald Metzler}
\affiliation{
\institution{Google Research}
}
\email{metzler@google.com}

\author{Cliff Brunk}
\affiliation{
\institution{Google Research}
}
\email{cliffbrunk@google.com}

\author{Andrew Tomkins}
\affiliation{
\institution{Google Research}
}
\email{tomkins@google.com}

\date{}

\begin{document}
\begin{abstract}
Large generative language models such as GPT-2 are well-known for their ability to generate text as well as their utility in \emph{supervised} downstream tasks via fine-tuning.
Our work is twofold: firstly we demonstrate via human evaluation that classifiers trained to discriminate between human and machine-generated text emerge as \emph{unsupervised} predictors of ``page quality'', able to detect low quality content without any training. This enables fast bootstrapping of quality indicators in a low-resource setting. Secondly, curious to understand the prevalence and nature of low quality pages in the wild, we conduct extensive qualitative and quantitative analysis over 500 million web articles, making this the largest-scale study ever conducted on the topic.
\end{abstract}

\maketitle
\pagestyle{plain} 
\section{Introduction}
The application of large neural language models for text generation has received a great deal of attention, from both the research community and the popular press \cite{radford2019language,radford2018improving,zellers2019defending,keskar2019ctrl,dathathri2019plug,solaiman2019release,seabrook2019}. Many have raised concerns about the potential dangers of neural text generators in the wild, owing largely to their ability to produce human-looking text at scale.

Classifiers trained to discriminate between human and machine-generated text have recently been employed to monitor the presence of machine-generated text on the web \cite{solaiman2019release}. Little work, however, has been done in in applying these classifiers for other uses, despite their attractive property of requiring no labels - only a corpus of human text and a generative model. In this work, we show through rigorous human evaluation that off-the-shelf human vs. machine discriminators serve as powerful classifiers of page quality. That is, texts that appear machine-generated tend to be incoherent or unintelligible. To understand the presence of low page quality in the wild, we apply the classifiers to a sample of half a billion English webpages. We analyze the results both qualitatively and quantitatively, providing breakdowns across dimensions such as time and topic distribution. We use two state-of-the-art methods for detecting machine-generated text: OpenAI's RoBERTa-based GPT-2 detector \cite{liu2019roberta,solaiman2019release} and GLTR \cite{gehrmann2019gltr}. These models are trained to distinguish GPT-2-generated text from human text. The goal of this work is not to improve detection modeling but to demonstrate the effectiveness of existing detection approaches on surfacing low quality pages on the web.

A webpage's quality is a function of many factors including but not limited to the reputability of the domain, its incoming or outgoing hyperlinks, the factual correctness of the content, the audio or video media present, and notions purely around the textual content. In this work we focus solely on \emph{linguistic} or \emph{language} quality (which we will define more precisely later) and we hereafter use the terms ``page quality'' and ``language quality'' interchangeably.

\paragraph{Our Contributions.} The contributions of our work can be summarized as follows:
\begin{itemize}
    \item We demonstrate through human evaluation that existing detectors of machine-generated text are effective at predicting low quality pages, outperforming, quite surprisingly, supervised spam classifiers. To our knowledge, this is the first use of machine detection for a different NLP task.
    \item Using half a billion webpages, we conduct the largest application of the detection models in the wild.
    \item We quantify the low quality pages that are surfaced by our detector models. We perform extensive analysis, breaking them down by attributes such as document length, age, and topic.
    \item We qualitatively characterize and categorize the nature of the low quality documents. We find traces of essay generation farms, machine translated text, keyword optimizations, and Not-Safe-For-Work (NSFW) content.
\end{itemize}
\begin{figure*}[!ht]
\begin{subfigure}{0.32\textwidth}
\large
\centering\noindent\fbox{%
   \tt \parbox{0.9\linewidth}{%
   \textbf{When she not eating dinner weight loss was indulged, she said Since you are looking for Lu, you let them in. He said that the commander of the evening primrose oil appetite Ranking how to lose weight by swimming laps suppressant Guardians army is the emperors family. The Recommended maximum fat burner unbiased weight loss supplement reviews shop guy promised to go out and move the two jars of bamboo sake into it. This is definitely not will probiotic help lose weight a small Now You Can Buy moringa tea lose weight High Potency chiropractic weight loss amount.}
    }}
\label{low_lq}
\caption{Low LQ}
\end{subfigure}
\begin{subfigure}{0.32\textwidth}
\large
\centering\noindent\fbox{%
   \tt \parbox{0.9\linewidth}{%
\textbf{More pics already!! I keep checking in looking for updates!! My wife gets on to me all the time for buying projects. I love doing restores. A couple years ago, I restored a 35 year old jon boat and turned it into a crappie fishing machine! My problem is, I never keep my projects. I enjoy the restore more than using the item so after I get done, I typically sell and buy a new project. Don't have any current projects but Im looking.}
    }}
\caption{Medium LQ}
\label{mid_lq}
\end{subfigure}
\begin{subfigure}{0.32\textwidth}
\large
\centering\noindent\fbox{%
   \tt \parbox{0.9\linewidth}{%
\textbf{For the past week, my instagram feed has been pushing a sponsored video of a grinning woman in a sheer pink skirt stepping into traffic and spinning in circles at a busy intersection behind the Grand Palais. One of the hashtags on the post is \#FrenchGirlStyle. I cross this intersection all the time. Trust me, there's nothing even remotely French about twirling around in traffic. Parisian drivers are not patient and they do not suffer fools gladly. Also, Parisian women don't grin. (They do smile quite warmly, contrary to popular belief, but only during personal interactions.)}
    }}
\caption{High LQ}
\label{high_lq}
\end{subfigure}
\caption{Low, medium, and high language quality examples, as deemed by the GPT-2 detector and both human raters. Texts were minimally modified to protect the identity of the author.}
\label{fig:lq_sample_text}
\end{figure*}
\section{Related Work}
In this section, we briefly review work on text generation, human vs. machine detection, socially good and bad uses of neural generative models, and linguistic text quality.

\paragraph{Generative Neural Language Models.} Neural text generation has attracted intense attention in recent years, largely owing to its ability to generate realistic text. This has been popularized by the GPT \cite{radford2018improving}, GPT-2 \cite{radford2019language}, and GPT-3 \cite{brown2020language} models, which demonstrate that pre-training language models on large corpora enables not only superior downstream performance but can result in high quality text generation. Subsequently, \cite{keskar2019ctrl} and \cite{zellers2019defending} proposed CTRL and Grover respectively, which focused largely on text generation conditioned on article metadata. These models are auto-regressive and the sampling strategy used significantly affects generation quality \cite{tay2020reverse}. Sampling methods include naive sampling from the full next-token softmax distribution, selecting the arg-max only, sampling from the top-$k$ scoring tokens \cite{fan2018hierarchical} or the nucleus / top-$p$ head of the distribution \cite{holtzman2019curious}, and various flavors of beam search \cite{vijayakumar2016diverse,kool2019stochastic}. Top-$k$ and top-$p$ are commonly used in practice. Alternatives to auto-regressive models have been proposed.

\paragraph{Detection Models.}
Models detecting machine-generated text have also garnered interest in recent years. Early work in this area focused on feature-engineered models based on, for instance, $n$-gram or bag-of-words \cite{grechnikov2009detection,badaskar2008identifying}. Recent work has focused on leveraging pre-trained language models, e.g. by exploiting features obtained from language model outputs. GLTR \cite{gehrmann2019gltr} utilizes top-$k$ token rank features obtained from conditioning a language model on the input text. GLTR can also be used in a zero-shot setting where the probabilities assigned may be used as detectors without any training.

Generative models, such as Grover \cite{welleck2019neural} and GPT-2, can also be used for sequence-level detection. This is in similar spirit to the standard fine-tuning methodology \cite{radford2018improving,devlin2018bert}. Additionally, \cite{bakhtin2019real} proposes training an energy-based model that learns to rank a sequence of human text over the top-$k$ machine generations, primed on some amount of human text at either end of the sequence. Their model achieves strong results when train-test distributions are similar but struggles to generalize when there is a mismatch.

Notably, a recent study \cite{tay2020reverse} proposes the new detection task of predicting the source of a generator model (i.e. the generator model architecture and hyper-parameter settings) from samples of generated text. The authors show that the task can be quite easy in certain cases.

\paragraph{Misuse.}
There are a number of potential ways that neural generative models can be misused, the most direct of which is in authoring content with malicious intent. OpenAI carried out a study to characterize the usage and detection of GPT-2 in the wild \cite{solaiman2019release}. The study reports that their ``threat monitoring did not find evidence of GPT-2 direct misuse in publicly-accessible forums but [they] did see evidence of discussion of misuse'' and that ``several governments have experimented with GPT-2 and other language models''. While they provide some insight into how and by whom GPT-2 can be used, their findings are mostly qualitative and high-level in nature.

\paragraph{Generation for Good.}
Learning to detect machine text seems to imply that such text is generally considered to be bad or harmful. This is not always the case. Machine-generated text has a wide range of highly useful applications, like grammar correction \cite{alikaniotis2019unreasonable}. Furthermore, the use of neural machine translation \cite{wu2016google} can be regarded as a positive move towards global accessibility. There have also been reports of positive GPT-2 usage, such as gaming \cite{NickWalton}, programming assistance \cite{TabNine}, writing assistance \cite{huggingface}, and poetry generation \cite{GwernBranwen}. 

\paragraph{Text quality.}
We highlight some prior work on text quality, which can be roughly divided into two categories: (1) classifying the linguistic quality of human-written content, and (2) assessing and improving the quality and diversity of neural text generations. For human-written text, \cite{arapakis-etal-2016-linguistic} characterizes the constituents of high editorial quality among news articles while \cite{mesgar2018neural} proposes a model that captures local coherence between sentences and demonstrates its use on readability assessment and essay scoring tasks. There have been significant efforts to catch low content quality, often referred to as spam, in large-scale web applications \citep{ntoulas2006detecting,bendersky2011quality,cormack2011efficient,metsis2006spam}. For machine generations, BLEU \citep{papineni2002bleu} is a well-established method for measuring quality but struggles at perceiving diversity, while Self-BLEU \citep{zhu2018texygen} captures diversity but struggles with quality. Constructing measures that capture both diversity and quality is an active area of research in the community \citep{alihosseini-etal-2019-jointly}.
\section{Experiments and Results} This section outlines our experimental setup and results.

\begin{figure}[]
    \small
    \centering
    \includegraphics[width=1.0\linewidth]{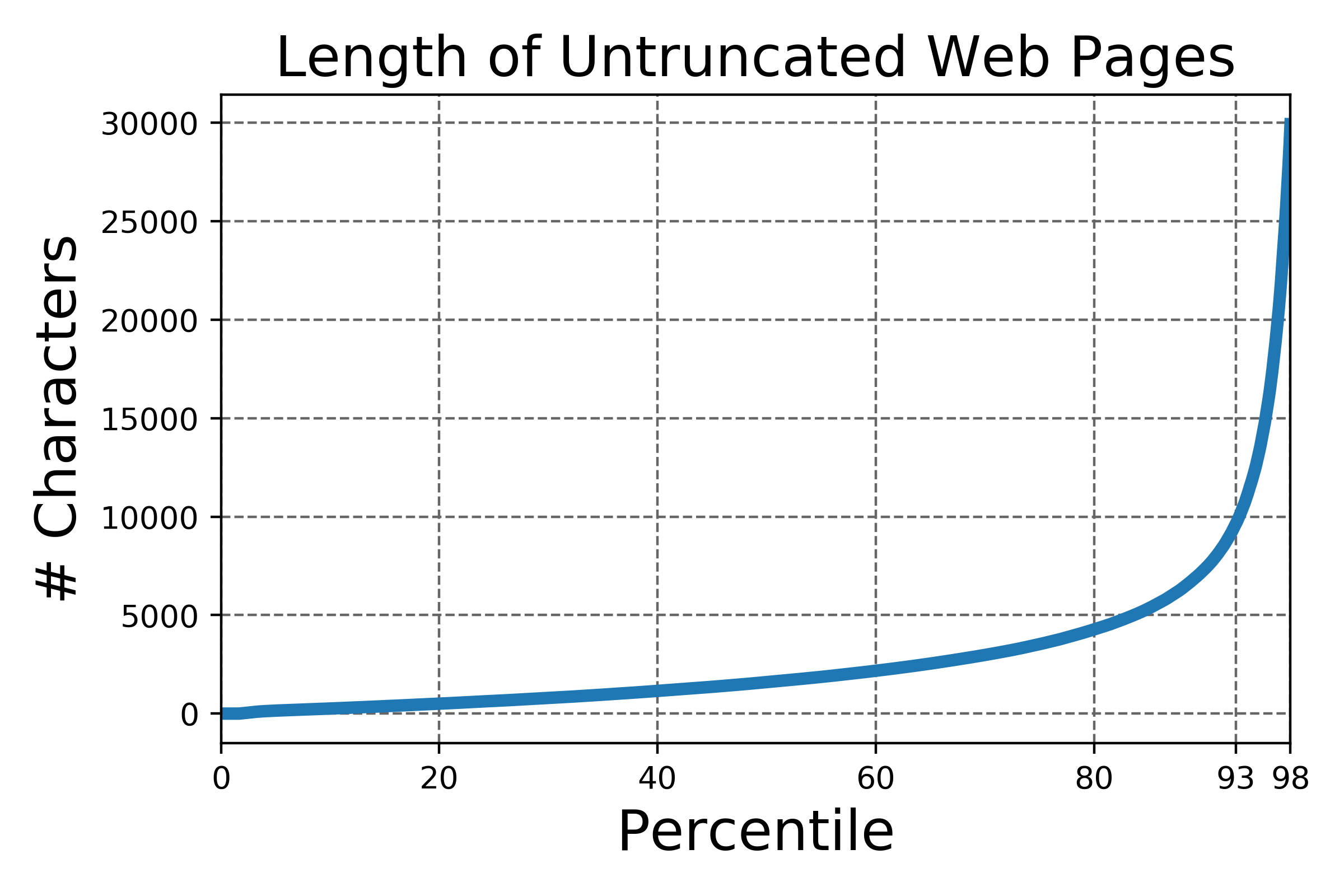}
    \vspace{-2mm}
    \caption{Percentile plot of document length for Web500M. About 93\% have fewer than 10k total characters, the length we truncate at.}
    \label{fig:character_hist}
\end{figure}

\begin{figure*}[!t]
\begin{minipage}{0.49\linewidth}
  \centering
    \includegraphics[width=0.9\linewidth]{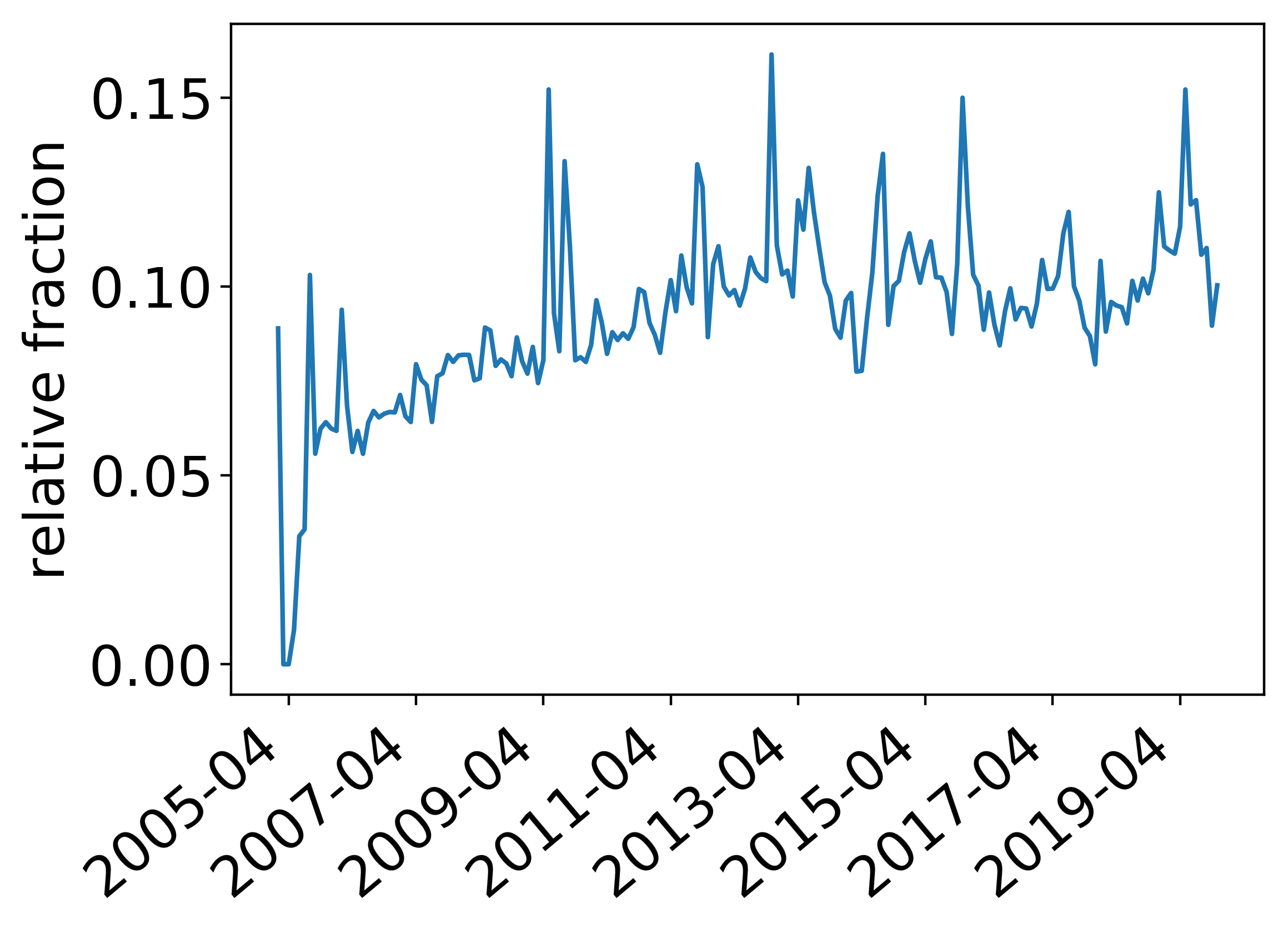}
    \subcaption{$\mathrm{score} > 0.5$}
\end{minipage}
\begin{minipage}{0.49\linewidth}
  \centering
    \includegraphics[width=0.9\linewidth]{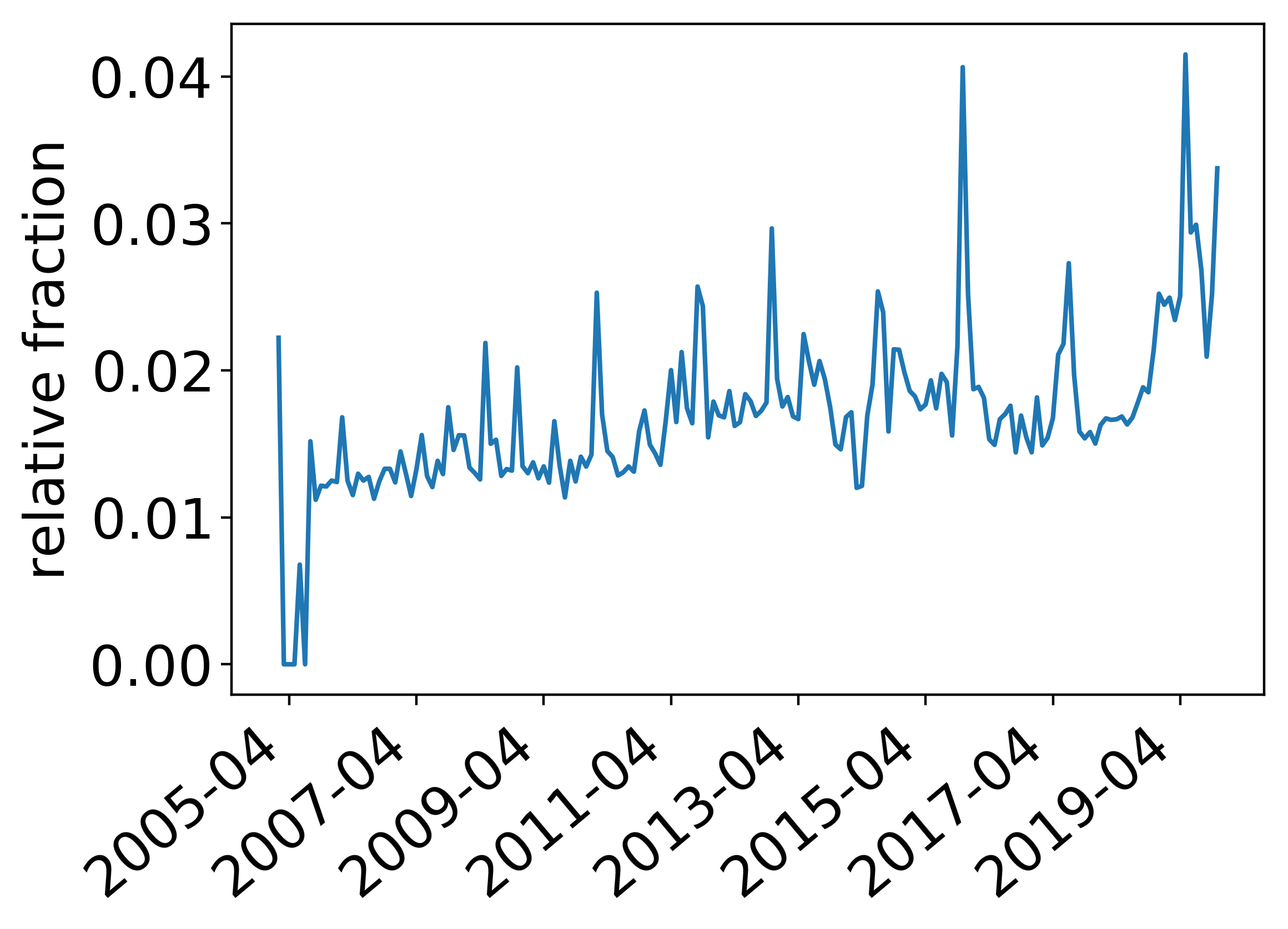}
    \subcaption{$\mathrm{score} > 0.8$}
\end{minipage}
\begin{minipage}{0.49\linewidth}
  \centering
    \includegraphics[width=0.9\linewidth]{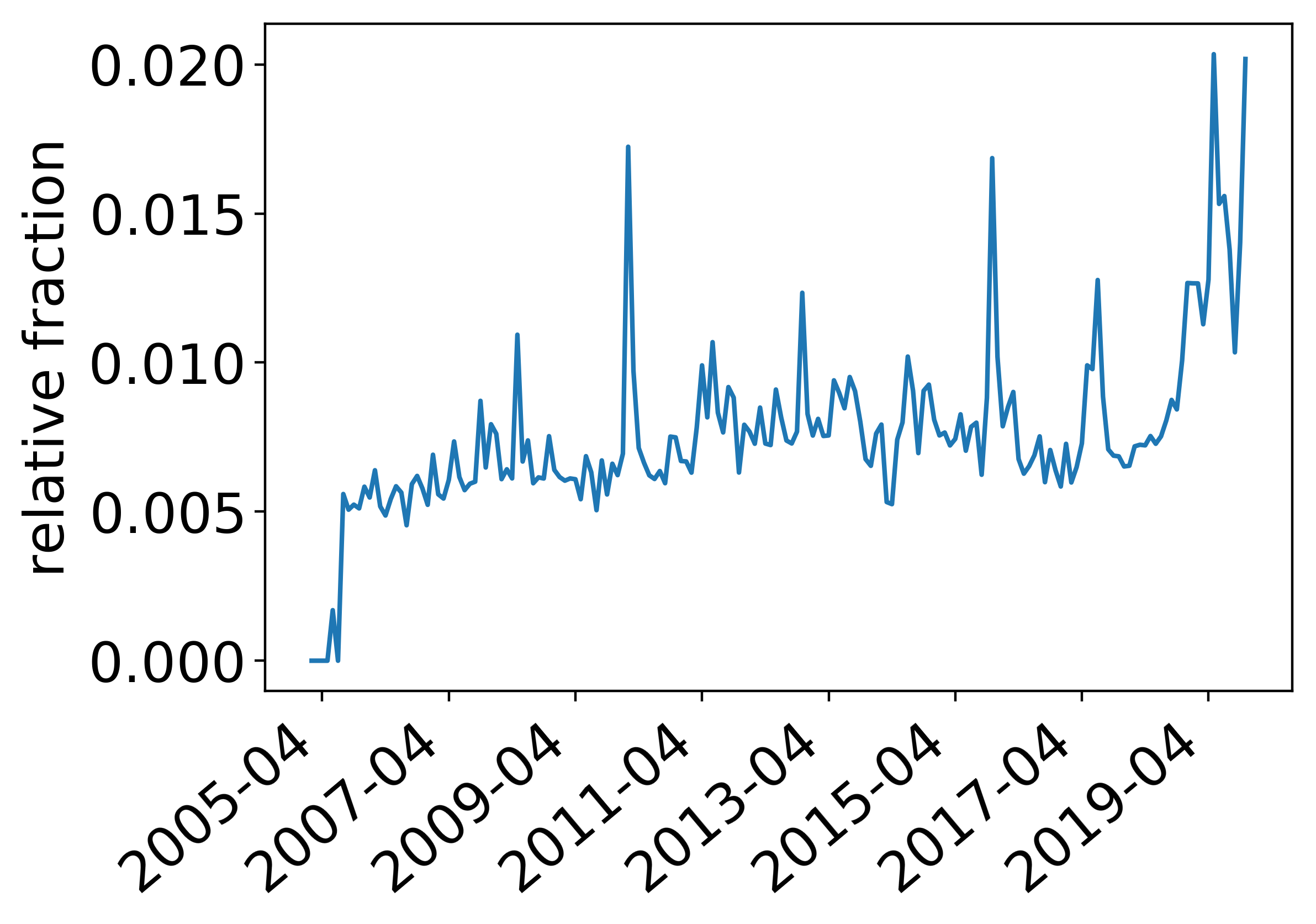}
    \subcaption{$\mathrm{score} > 0.9$}
\end{minipage}
\begin{minipage}{0.49\linewidth}
  \centering
    \includegraphics[width=0.9\linewidth]{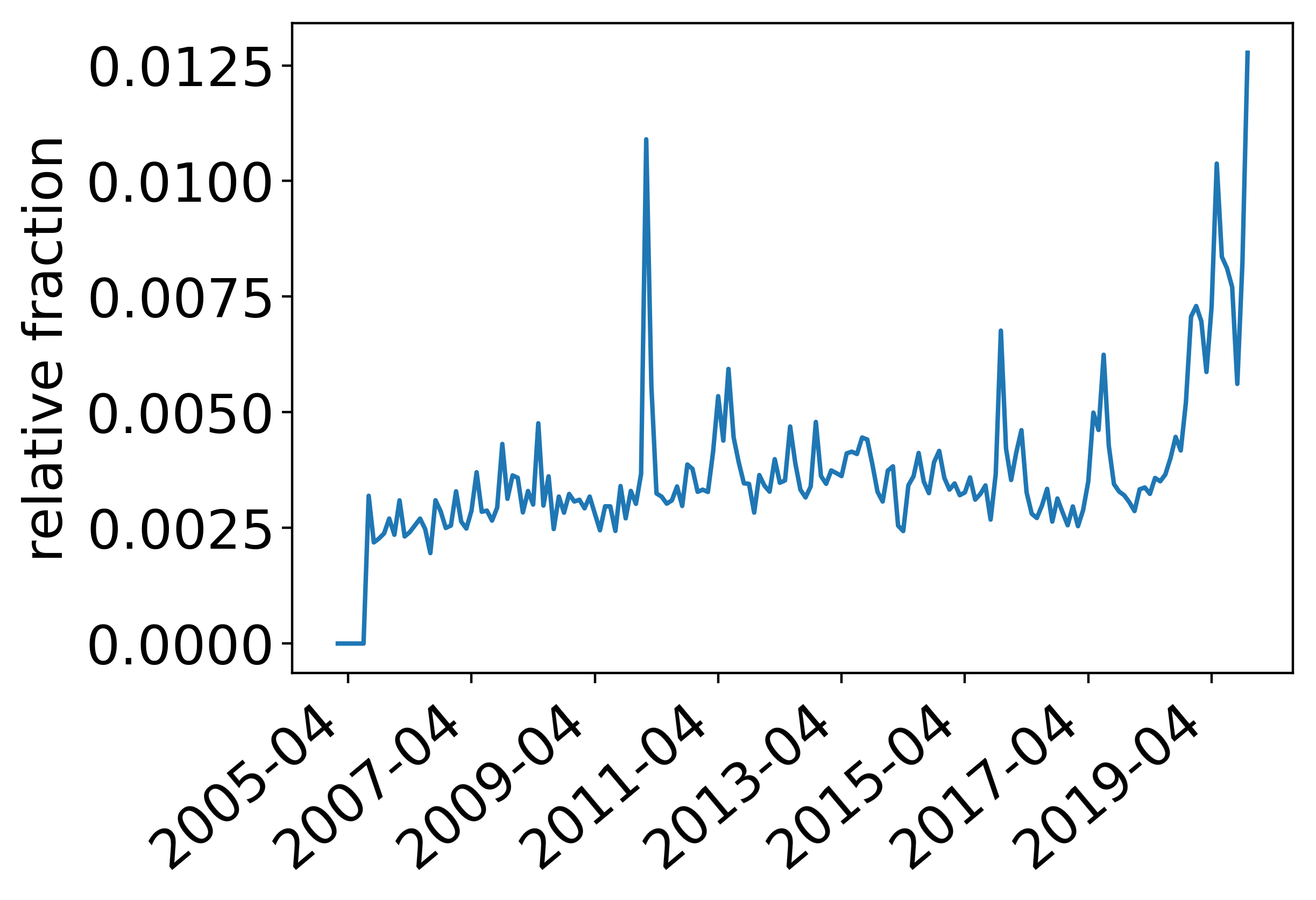}
    \subcaption{$\mathrm{score} > 0.95$}
\end{minipage}
\begin{minipage}{0.49\linewidth}
  \centering
    \includegraphics[width=0.9\linewidth]{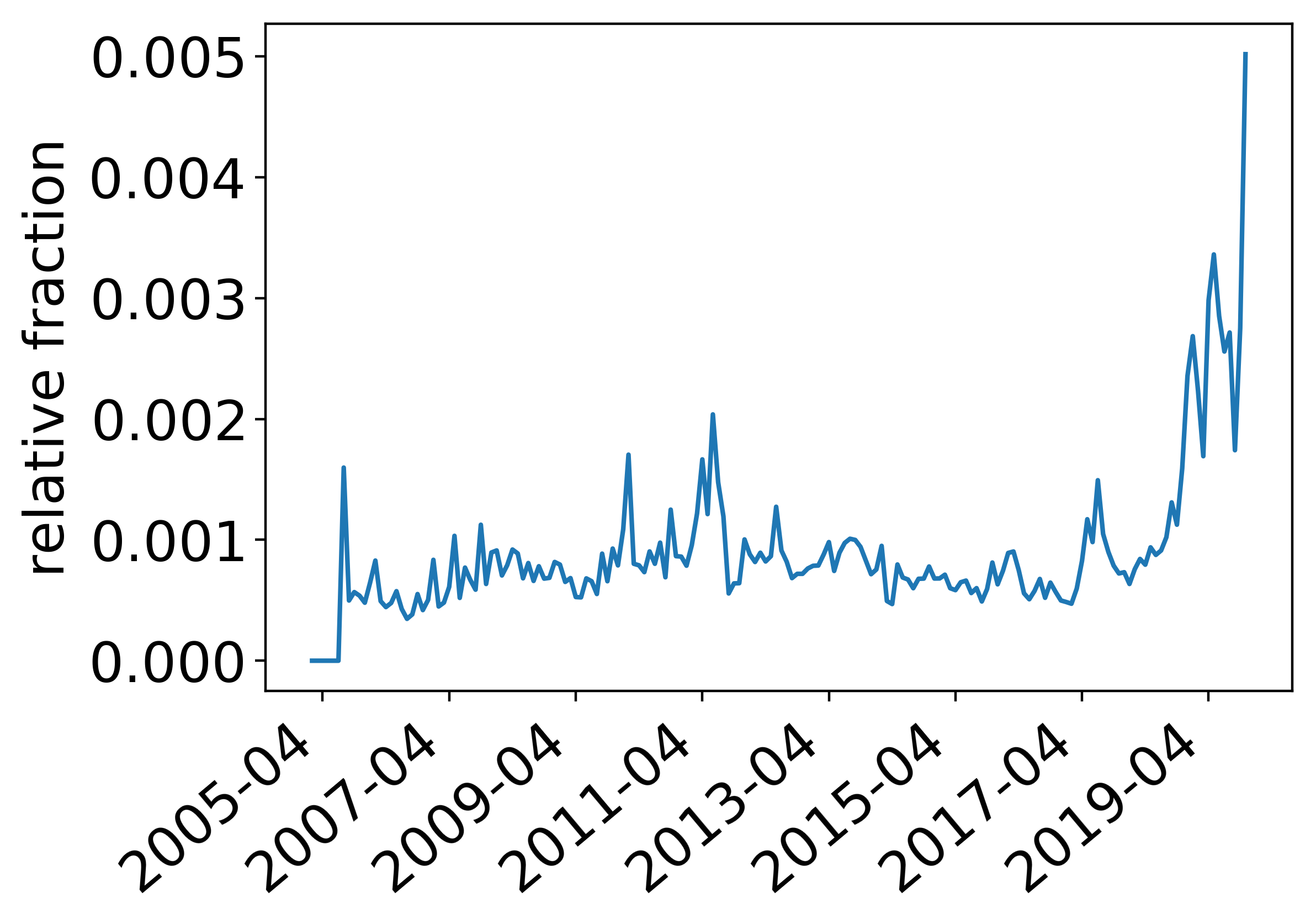}
    \subcaption{$\mathrm{score} > 0.99$}
\end{minipage}
\begin{minipage}{0.49\linewidth}
  \centering
    \includegraphics[width=0.9\linewidth]{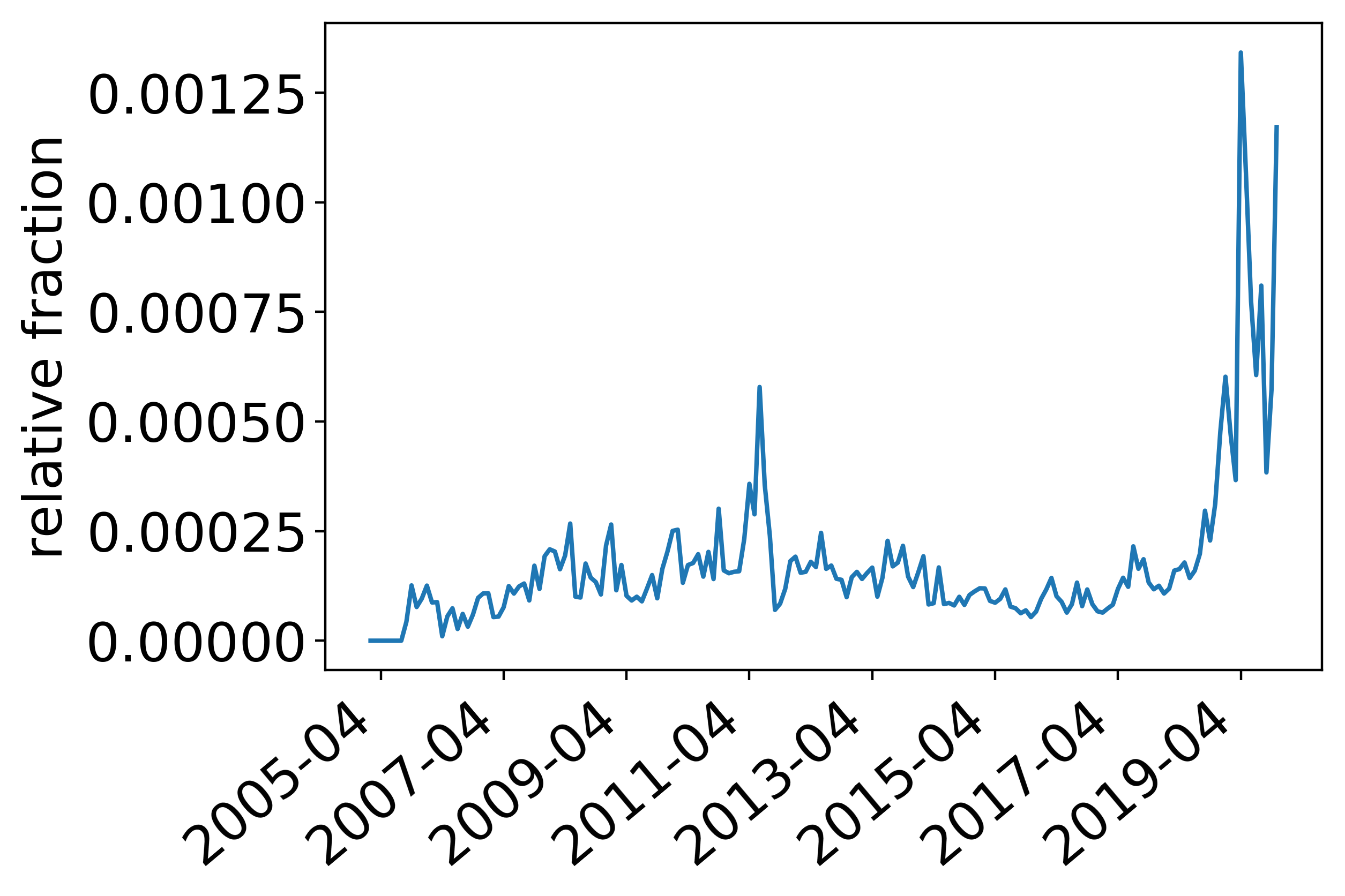}
    \subcaption{$\mathrm{score} > 0.999$}
\end{minipage}
    \caption{Relative fraction of documents over time at various OpenAI detector score threshold. We observe a burst in the fraction of low quality documents at the beginning of 2019.}
    \label{fig:temporal_openai}
\end{figure*}

\subsection{Datasets}
This section describes the datasets used in our experiments.
\begin{itemize}
\item \textbf{Web500M.} The core corpora used in our experiments consists of a random sample of 500 million English web documents obtained from the Common Crawl\footnote{\url{https://commoncrawl.org/}.}.

\item \textbf{GPT-2-Output.} This public dataset (\ref{openaiurl}) consists of the WebText test split and its GPT-2 generations under 8 different settings (2 sampling methods for each of 4 model sizes). The set is divided into a train, test, and validation split consisting of 250k, 5k, and 5k examples respectively.

\item \textbf{Grover-Output.} We generated $1.2$ million articles using pre-trained Grover-Base with a diverse range of sampling hyperparameters: top-$k$ sampling with $k$ ranging from $10$ to $100$ in steps of $10$ and top-$p$ sampling with
$p$ ranging from $0.65$ to $0.90$ in steps of $0.5$. The model was conditioned on only the title field of articles from the publicly available CNN/DailyMail dataset \cite{hermann2015teaching}. We used the public Grover source code\footnote{\url{https://github.com/rowanz/grover}} for generating this data.
\end{itemize}

In order to bound document detection times, we truncate each document in all datasets to its first 10k characters. As depicted in Figure \ref{fig:character_hist}, about 93\% of Web500M documents have fewer than 10k characters and are thus unaffected by the truncation.


\begin{figure*}[h]
\centering
\begin{minipage}{0.30\linewidth}
  \centering
\includegraphics[width=1.0\linewidth]{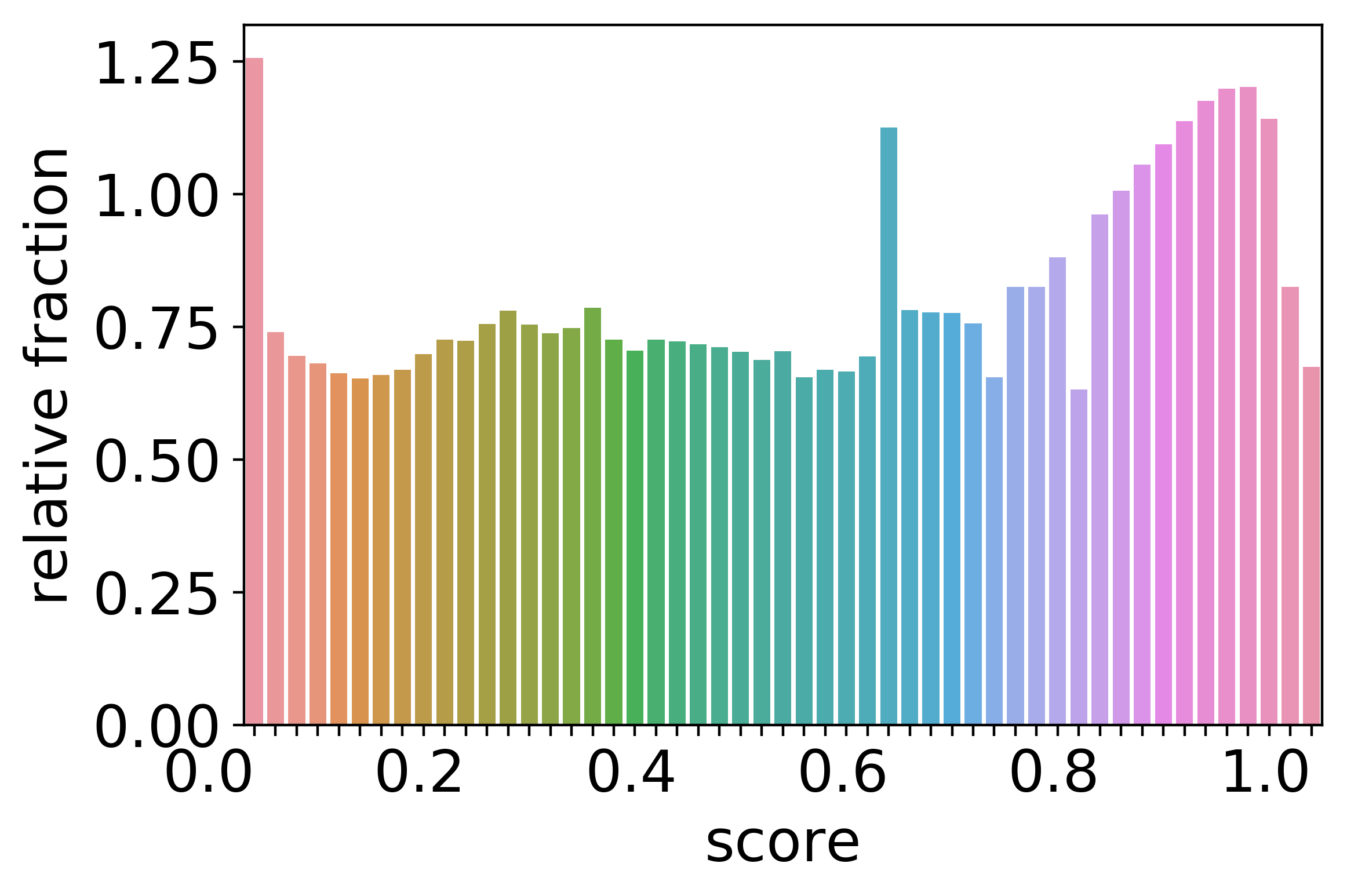}
\subcaption{News}
\label{topic1}
\end{minipage}
\begin{minipage}{0.30\linewidth}
  \centering
\includegraphics[width=1.0\linewidth]{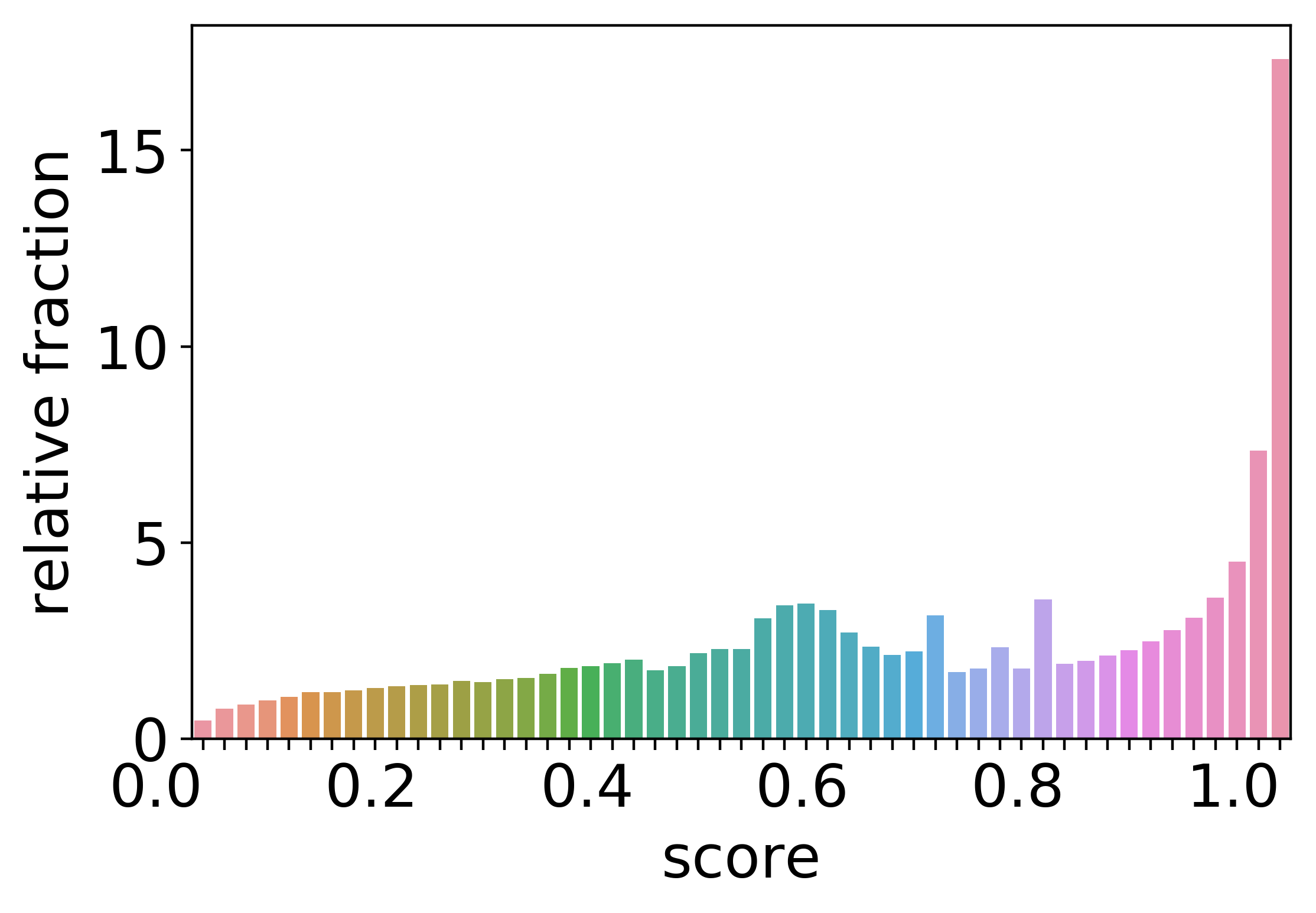}
\subcaption{Adult}
\label{topic2}
\end{minipage}
\begin{minipage}{0.30\linewidth}
  \centering
\includegraphics[width=1.0\linewidth]{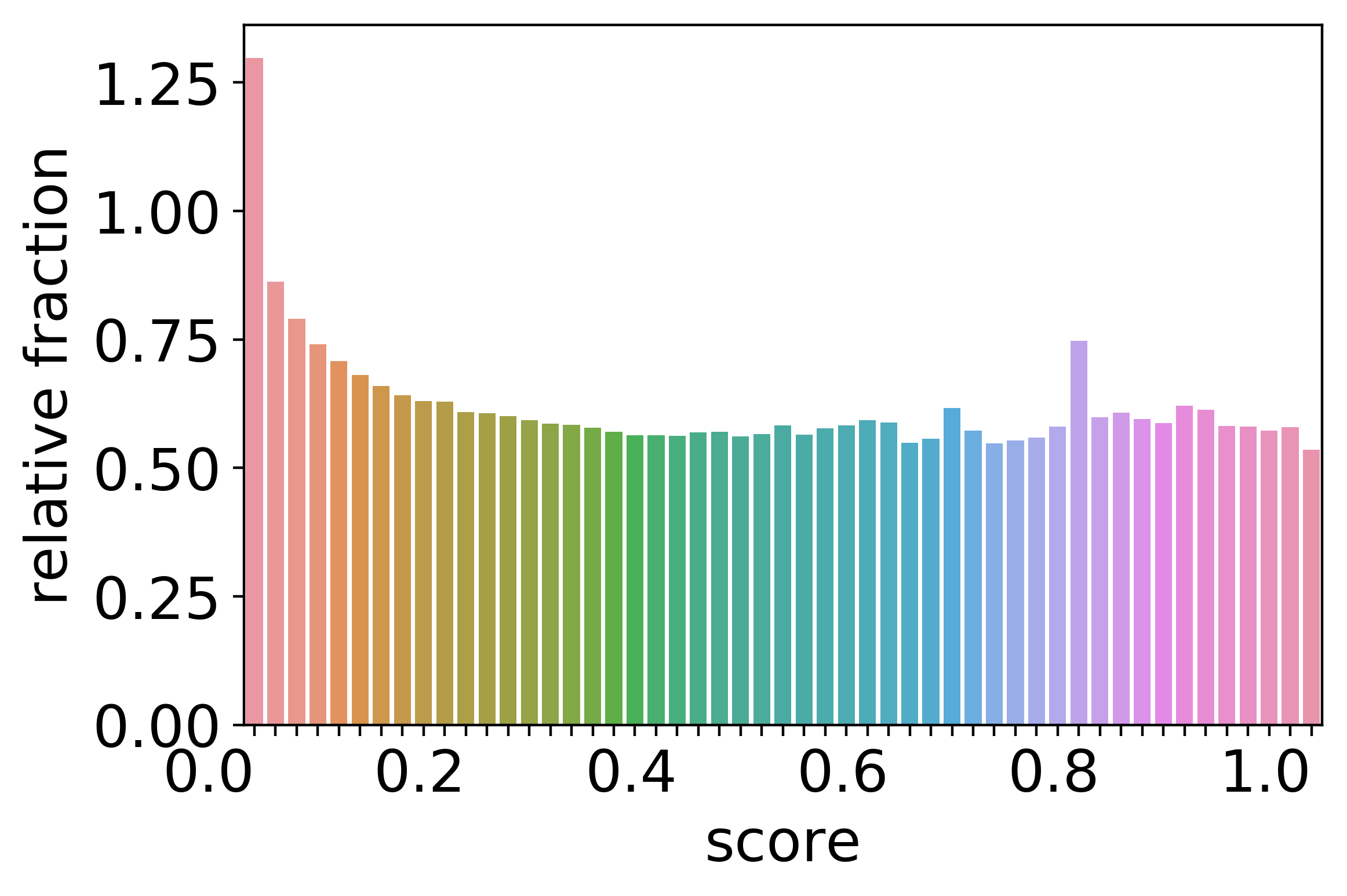}
\subcaption{Law/Government}
\label{topic3}
\end{minipage}
\centering
\begin{minipage}{0.30\linewidth}
  \centering
\includegraphics[width=1.0\linewidth]{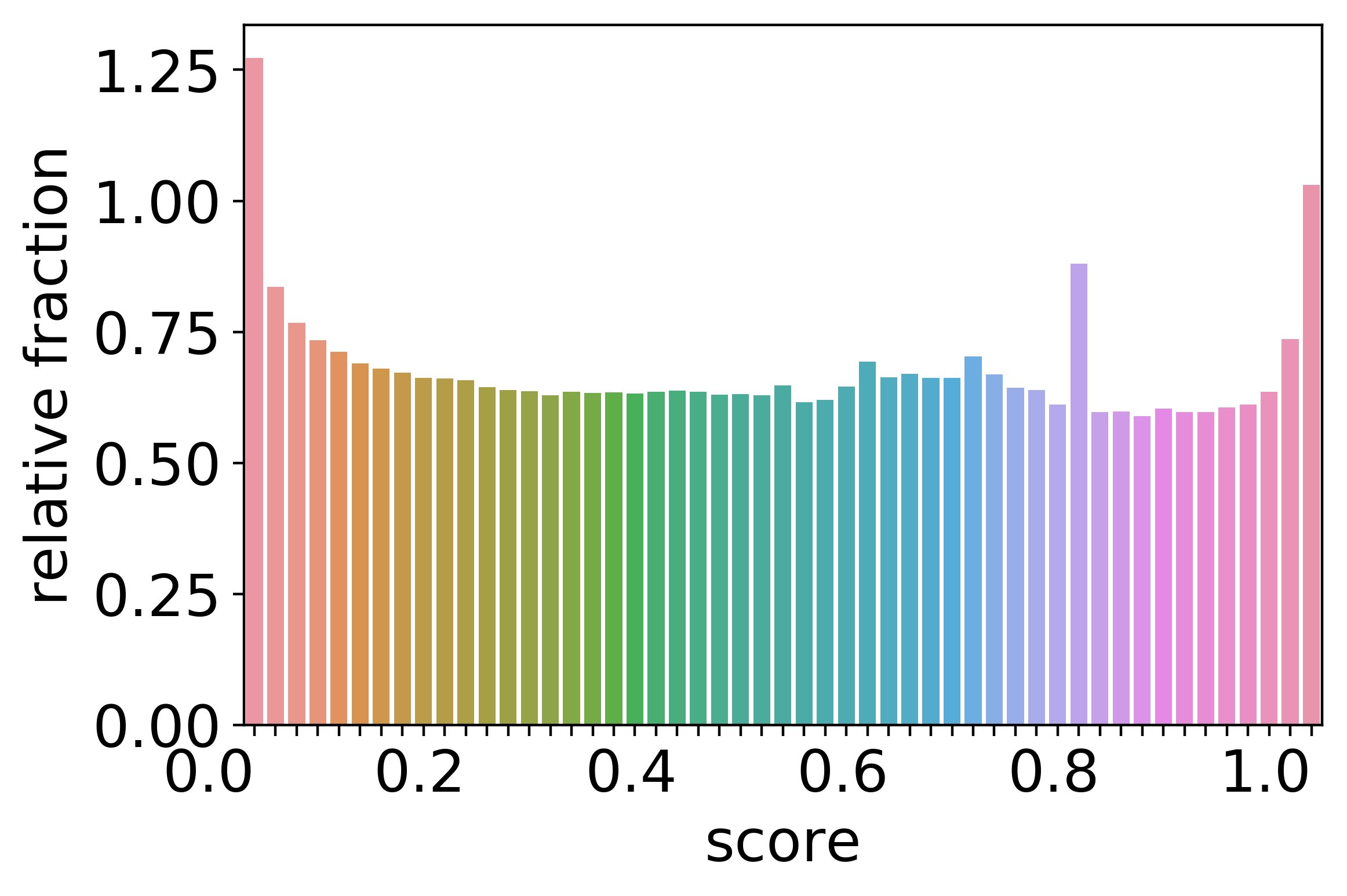}
\subcaption{People and Society}
\label{topic4}
\end{minipage}
\begin{minipage}{0.30\linewidth}
  \centering
\includegraphics[width=1.0\linewidth]{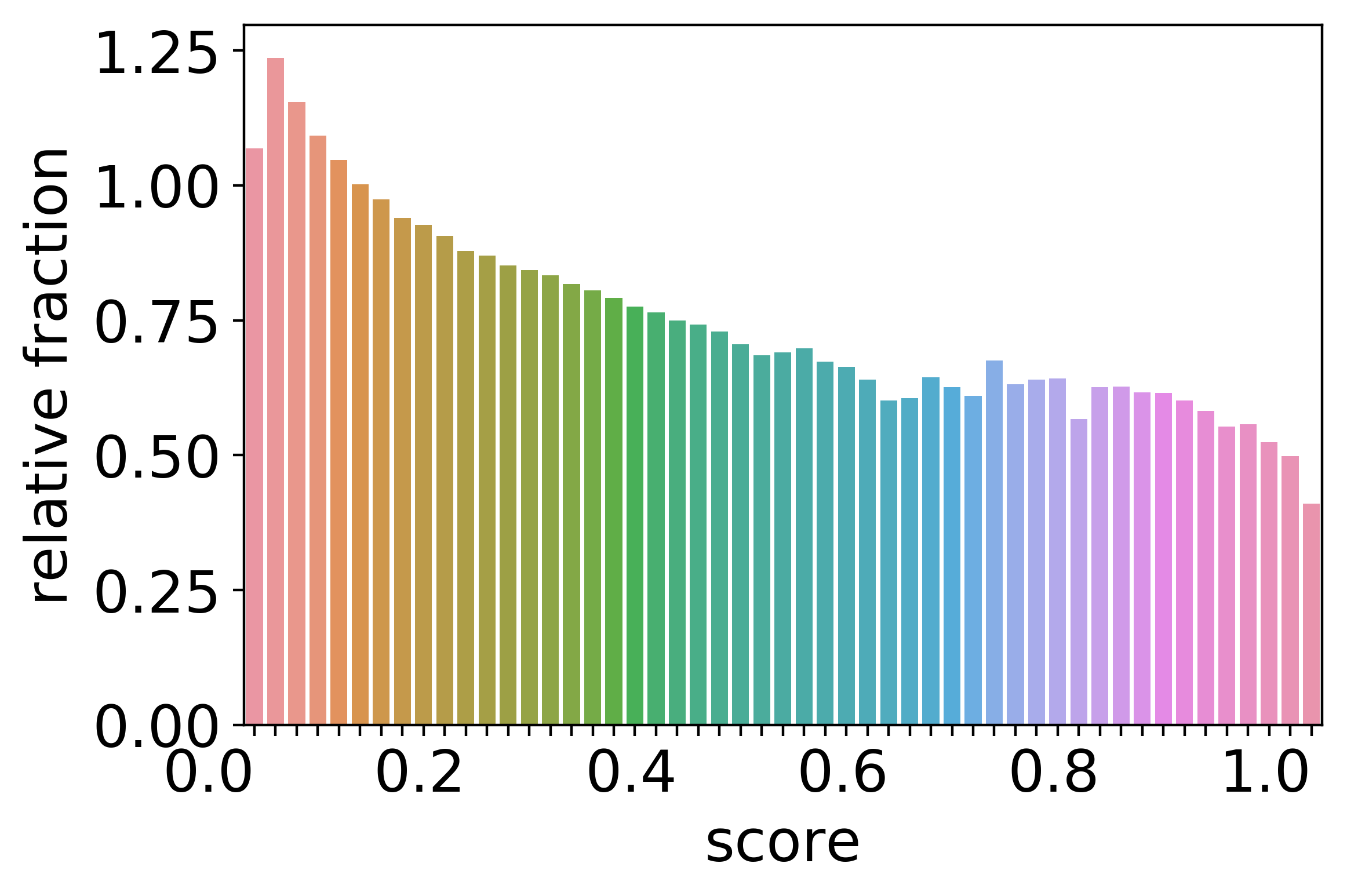}
\subcaption{Science}
\label{topic5}
\end{minipage}
\begin{minipage}{0.30\linewidth}
  \centering
\includegraphics[width=1.0\linewidth]{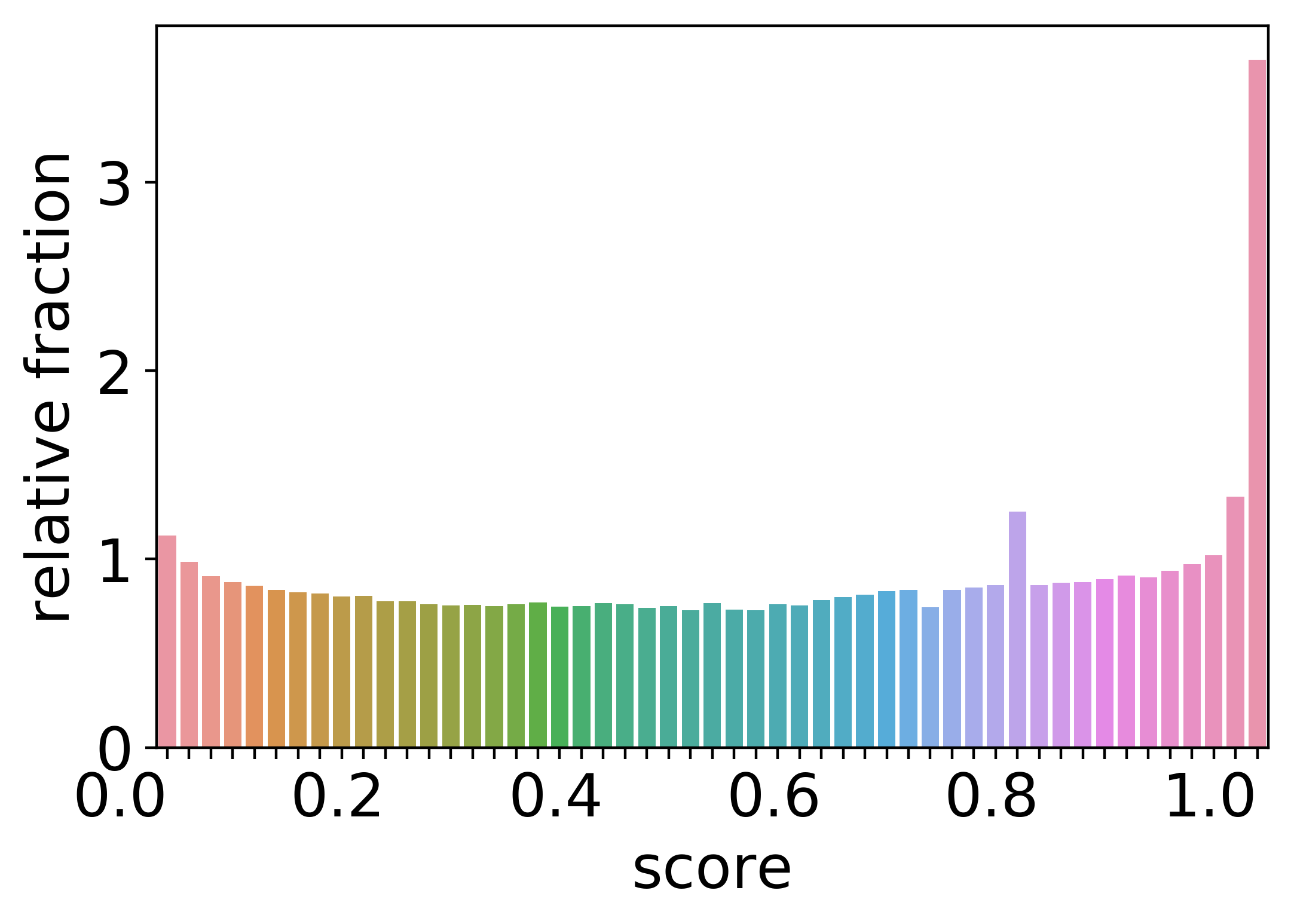}
\subcaption{Books/Literature}
\label{topic6}
\end{minipage}

\begin{minipage}{0.30\linewidth}
  \centering
\includegraphics[width=1.0\linewidth]{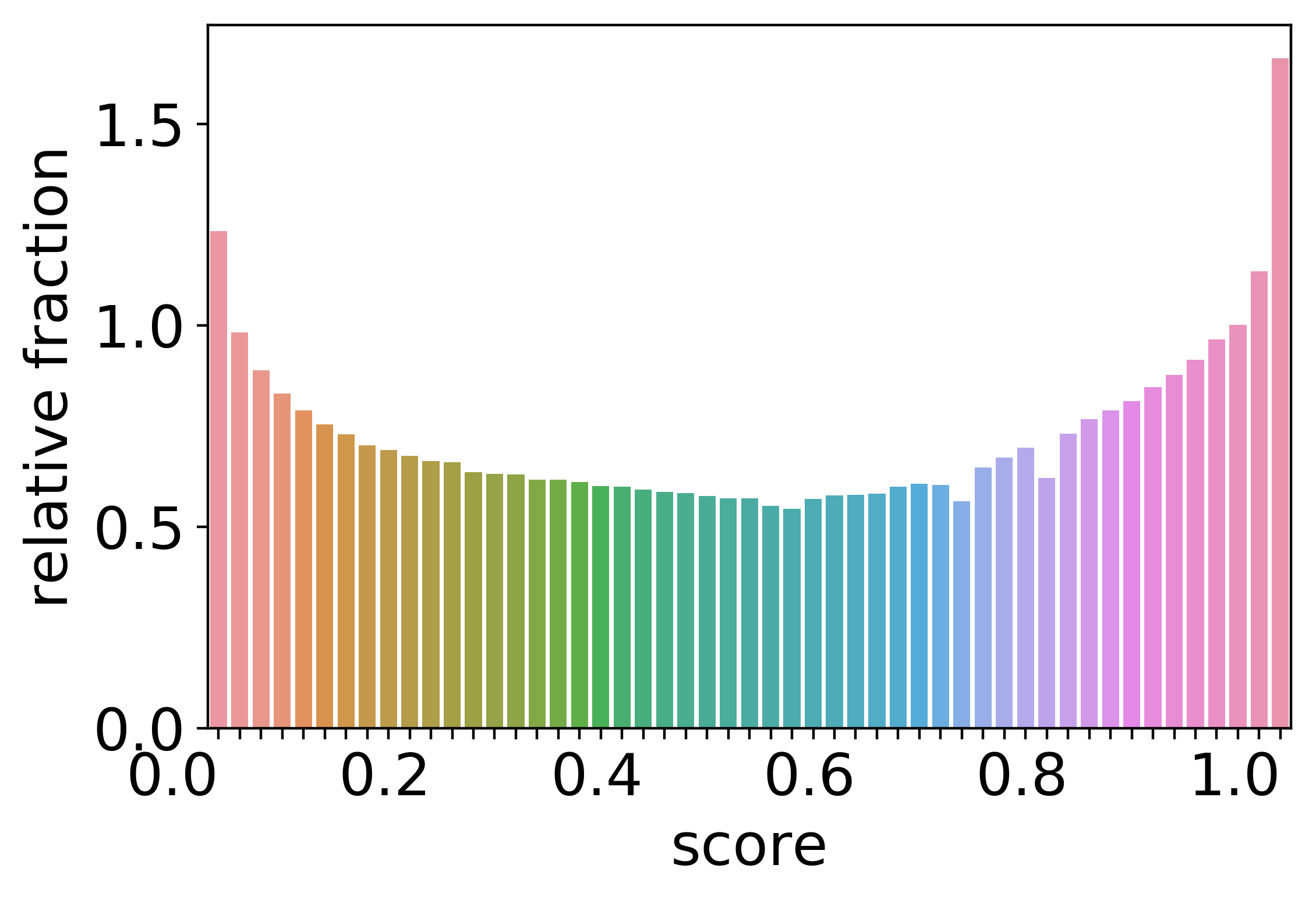}
\subcaption{Health}
\label{topic4}
\end{minipage}
\begin{minipage}{0.30\linewidth}
  \centering
\includegraphics[width=1.0\linewidth]{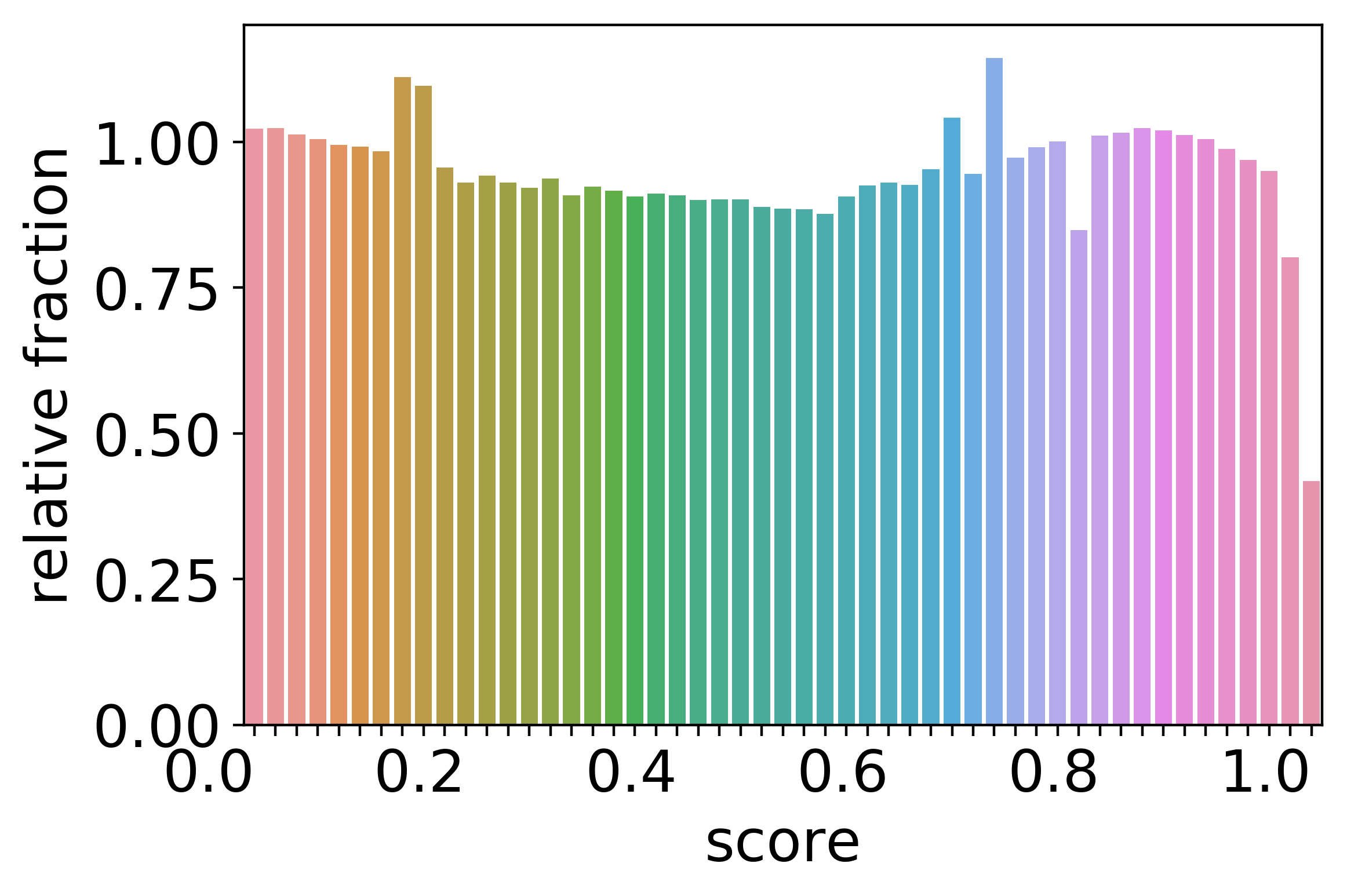}
\subcaption{Food}
\label{topic5}
\end{minipage}
\begin{minipage}{0.30\linewidth}
  \centering
\includegraphics[width=1.0\linewidth]{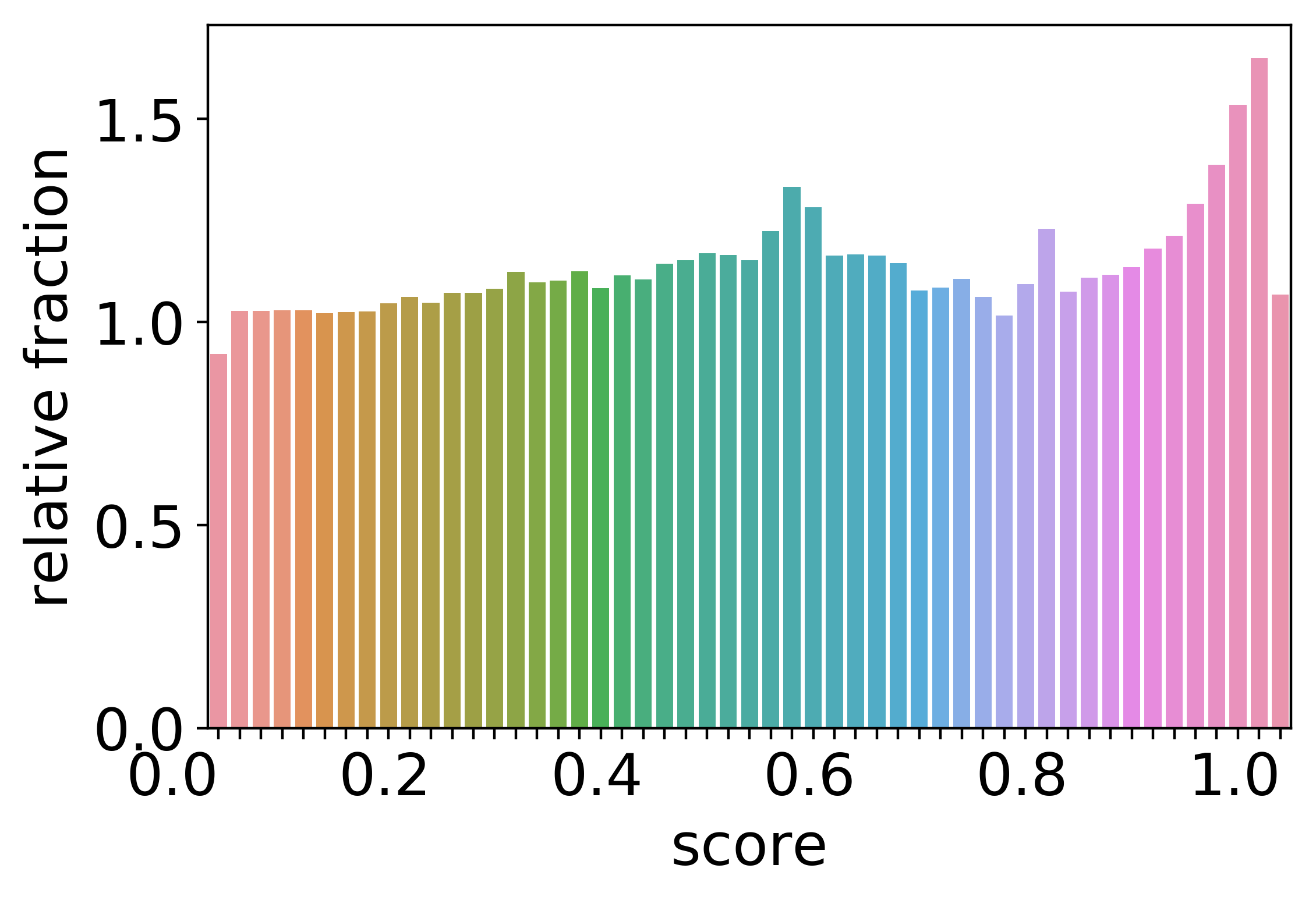}
\subcaption{Games}
\label{topic6}
\end{minipage}

\caption{Relative fraction of OpenAI detector scores for different topics. }
\label{topical_histograms}
\end{figure*}

\subsection{Detectors}
Our experiments utilize two recent detection methods, which we implement using the Tensorflow API of HuggingFace's Transformers library \cite{abadi2016tensorflow,Wolf2019HuggingFacesTS}.

To understand the significance of our two detectors as unsupervised predictors of page quality, we compare against a baseline that was trained explicitly on the spam classification task using the Scikit-Learn~\cite{scikit-learn} python package. Since page quality and "spaminess" are strongly related, we expect that a classifier trained to detect the latter will transfer well for assessing the former.

\begin{itemize}
\item \textbf{OpenAI's GPT-2 Detector.} We use OpenAI's publically available\footnote{\label{openaiurl}\url{https://github.com/openai/gpt-2-output-dataset}} detection model, a RoBERTa-large (356 million parameters) that was fine-tuned on GPT-2 generations using a mix of untruncated and top-$k$ 40 sampling. For each sequence no longer than 512 byte-level Byte-Pair-Encoding (BPE) tokens, the detector outputs the probability of that sequence being machine written. For longer texts, we use the score from the first 512-length sequence.

\item \textbf{GLTR.} We follow the method proposed in \citet{gehrmann2019gltr} but use the much larger GPT-2 XL model (1.5 billion parameters). For each token $T$ in the target text, we obtain a softmax distribution by conditioning the model on text surrounding $T$. Next, we compute the integer rank of $T$ in the sorted distribution.
We then construct a document-level histogram over the individual token ranks binned into top-$k$ buckets, where $k\in\{10,100,1000,|V|\}$ and $|V| = 50,257$ is the size of the token vocabulary. We train a logistic regression
on this single 4-dimensional feature using the train split of GPT2-Output. Concretely, we take 250k human-authored documents and 250k GPT-2-generated documents split uniformly across the 8 settings.

\item \textbf{Spam Baseline.}
We train a spam / not-spam classifier using the Enron Spam Email dataset~\cite{metsis2006spam}\footnote{\url{http://nlp.cs.aueb.gr/software_and_datasets/Enron-Spam/index.html}}. We construct train and test splits; train comprises $12875$ documents for each spam and not-spam (a.k.a. ham) class, while test similarly comprises $3219$ documents. Using the train split only, we learn a TF-IDF histogram featurizer using a vocabulary of $5000$ lowercase words. We then train a logistic regression classifier on top of the 5000-dimensional featurized training documents. The featurizer and classifier combination achieves $96.9\%$ accuracy on the test split. In the language quality evaluation described in the next section, we use the model's calibrated estimate of the probability of not-spam as its effective language score. In other words, if the model estimates that a document is not-spam with probability $0.2$, then its language quality score on this document is also $0.2$.
\end{itemize}


\large
\begin{table}[]
\centering
\begin{tabular}{l|c|c}
Method & Corr. & 95\% CI \\ \toprule
OpenAI & \textbf{0.740} & \textbf{[0.637, 0.822]} \\
\hline
Spam Baseline & 0.454 & [0.309, 0.582] \\
\midrule
\midrule
GLTR LR & \textbf{0.713} & \textbf{[0.569, 0.835]} \\              \hline
Spam Baseline & 0.495 & [0.316, 0.659] \\
\bottomrule
\end{tabular}

\caption{Pearson correlation between human and classifier LQ scores. We observe a large, statistically significant, positive correlation for both models, indicating that they are effective predictors of language quality. Furthermore, in both cases, the correlation is stronger than the baseline, which was trained with supervision for spam detection.}
\label{tab:lq_results}
\end{table}

\large
\begin{table}[]
\centering
\begin{tabular}{l|c|c}
& Cohen's kappa & 95\% CI \\ \toprule
GLTR LR    & 0.501                & [0.387, 0.611]                 \\ \hline
OpenAI & \textbf{0.604}               & \textbf{[0.474, 0.724]}                \\ \bottomrule
\end{tabular}
\caption{Cohen's kappa coefficient for inter-rater reliability. The two raters achieve high, statistically significant agreement on the four possible LQ categories (including ``Undefined'').}
\label{tab:cohen_kappa}
\end{table}

\begin{figure*}[!ht]
\centering
\begin{minipage}{0.32\linewidth}
  \centering
\includegraphics[width=1.0\linewidth]{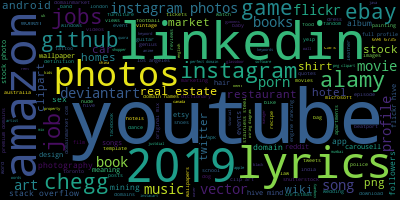}
\subcaption{$0.01<\mathrm{score}<0.1$.}
\label{wordclouds1}
\end{minipage}
\begin{minipage}{0.32\linewidth}
  \centering
\includegraphics[width=1.0\linewidth]{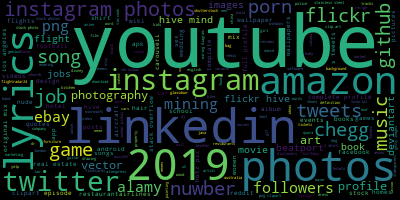}
\subcaption{$0.1<\mathrm{score}<0.2$}
\label{wordclouds2}
\end{minipage}
\centering
\begin{minipage}{0.32\linewidth}
  \centering
\includegraphics[width=1.0\linewidth]{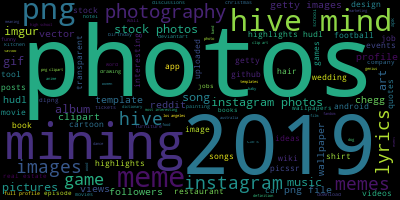}
\subcaption{$0.5<\mathrm{score}<0.6$}
\label{wordclouds3}
\end{minipage}
\begin{minipage}{0.32\linewidth}
  \centering
\includegraphics[width=1.0\linewidth]{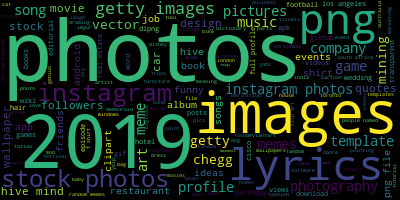}
\subcaption{$0.6<\mathrm{score}<0.7$}
\label{wordclouds4}
\end{minipage}
\centering
\begin{minipage}{0.32\linewidth}
  \centering
\includegraphics[width=1.0\linewidth]{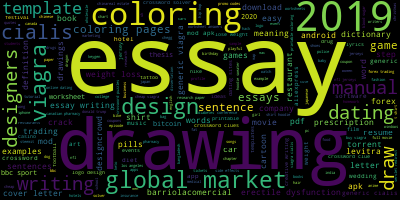}
\subcaption{$0.99<\mathrm{score}<1.0$}
\label{wordclouds5}
\end{minipage}
\begin{minipage}{0.32\linewidth}
  \centering
\includegraphics[width=1.0\linewidth]{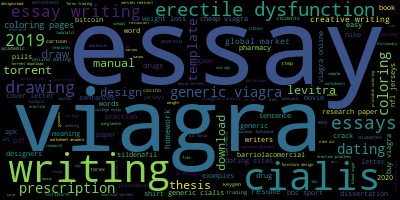}
\subcaption{$0.999<\mathrm{score}<1.0$}
\label{wordclouds6}
\end{minipage}
\caption{Word Cloud on Web500M for different ranges of OpenAI detector scores.}
\label{fig:wordcloud}
\end{figure*}

\subsection{Language Quality Evaluation}
As mentioned earlier, while we use ``page'' and ``language'' quality interchangeably, we are specifically interested in the quality of the language on the page. To that end, we define a language quality (LQ) score using the following criteria:
\begin{itemize}
    \item 0: Low LQ. Text is incomprehensible or logically inconsistent.
    \item 1: Medium LQ. Text is comprehensible but poorly written (frequent grammatical / syntactical errors).
    \item 2: High LQ. Text is comprehensible and reasonably well-written (infrequent grammatical / syntactical errors).
    \item Undefined: LQ is hard to assess for any reason.
\end{itemize}
We evaluate our machine vs. human classifiers and baseline using this criteria. To better assess language quality, we first filter Web500M by dropping all samples with fewer than 7.5k characters. Next, we define 3 buckets on the filtered Web500M corpus using percentiles of the classifier's P(machine-written) score: bottom = [0, 0.5], middle = [50, 50.5], top = [99.5, 100]. We sample 35 documents from each bucket for a total of 105 documents. Documents from the bottom, middle, and top buckets are assigned LQ scores of 2, 1, and 0 respectively. All documents are then rated by two human raters using the aforementioned criteria. Documents that at least one rater marked ``Undefined'' are dropped and all other documents are assigned a composite score that is the average of the raters' scores. Finally, we compute the Pearson correlation coefficient between the human and classifier's LQ scores along with a 95\% bootstrap confidence interval. To measure the inter-rater reliability, or degree of agreement between the two raters, we compute Cohen's kappa coefficient along with a 95\% bootstrap confidence interval.

Correlation and inter-rater reliability results are shown in Table~\ref{tab:lq_results} and Table~\ref{tab:cohen_kappa} respectively. Samples illustrating various language quality scores are shown in Figure~\ref{fig:lq_sample_text}. For both models, both the Pearson correlation and the inter-rater agreement is large and statistically significant, suggesting that documents with high P(machine-written) score tend to have low language quality. Furthermore, the models outperform the baseline, which was trained for spam detection in a supervised fashion. Machine authorship detection can thus be a powerful proxy for quality assessment. It requires no labeled examples - only a corpus of text to train on in a self-discriminating fashion. This is particularly valuable in applications where labeled data is scarce or where the distribution is too complex to sample well. For example, it is challenging to curate a labeled dataset representative of \textit{all} forms of low quality web content.

\subsection{Detector Performance}
\begin{figure*}[!t]
\centering
    \begin{minipage}{0.48\linewidth}
    \includegraphics[width=0.90\linewidth]{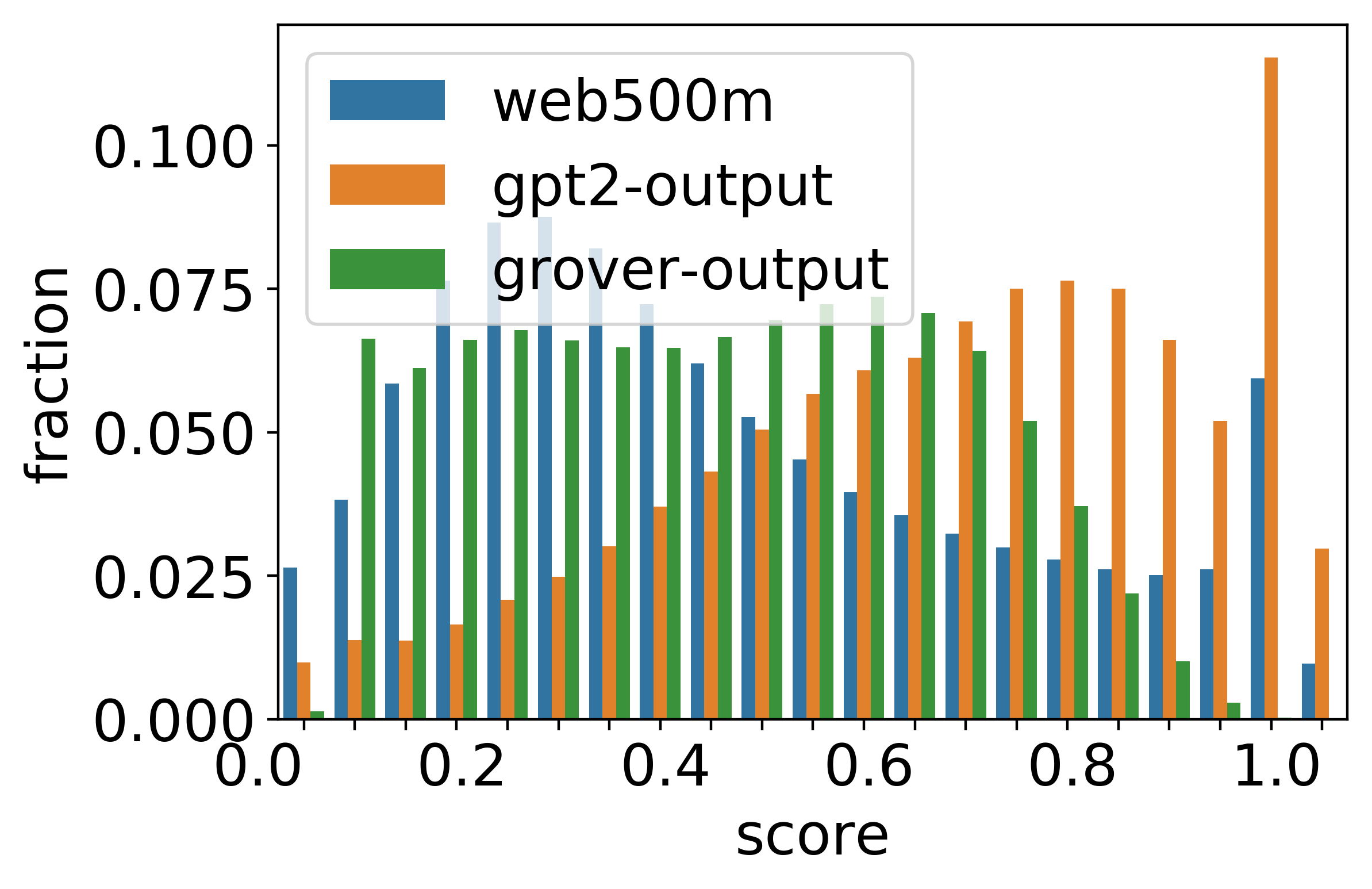}
    \subcaption{GLTR Logistic Regression}
    \label{fig:global_scores_gltr_lr}
    \end{minipage}%
    \begin{minipage}{0.48\linewidth}
    \includegraphics[width=0.90\linewidth]{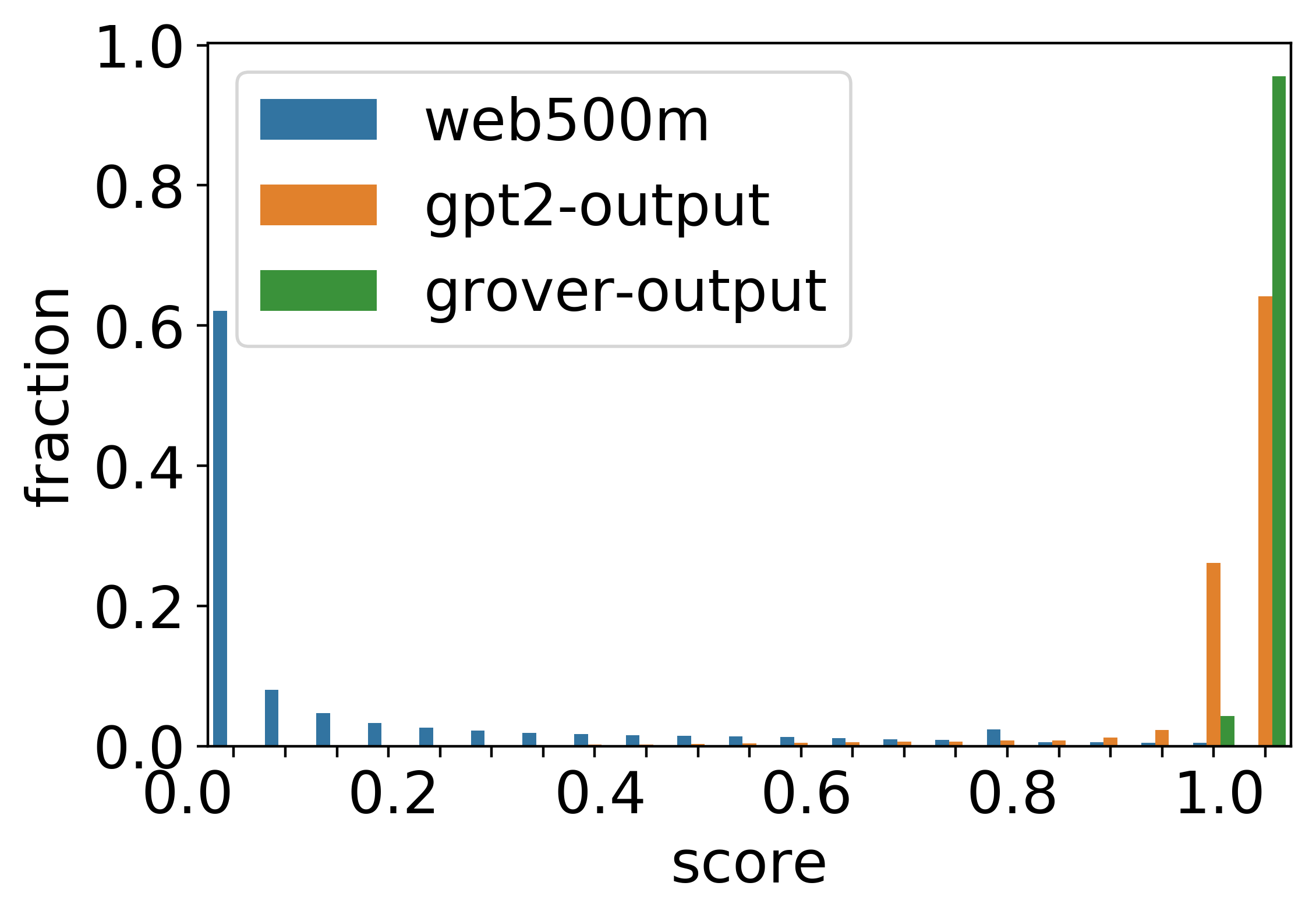}
    \subcaption{OpenAI Detector}
    \label{fig:global_scores_openai}
    \end{minipage}%
    \caption{Score distributions for GLTR and OpenAI detectors respectively. The OpenAI detector separates web and machine-generated documents much more cleanly than GLTR.}
    \label{fig:global_scores}
\end{figure*}

Table~\ref{tab:clf_accuracy} shows test accuracy of the detectors on both GPT-2 and Grover distributions. Contrary to our intuition, we find that the OpenAI detector generalizes to the Grover distribution better than the simpler GLTR detector. In Figure~\ref{fig:global_scores} we compare the distribution of the detector scores on the web corpus against that of machine-written texts.
Unlike the GLTR logistic regression detector, the OpenAI detector's score distribution is well-separated: scores are either either small or large. We focus on the OpenAI detector in subsequent analysis in light of its better predictive performance on machine vs. human discrimination as well as a higher correlation with human-rated LQ labels, as described in the previous section.

\large
\begin{table}[]
\centering
\begin{tabular}{l|c|c}
& GPT-2 & Grover \\ \toprule
GLTR LR & 75\%               & 80\%                 \\ \hline
OpenAI    & \textbf{85\%}                & \textbf{82\%}                 \\ \bottomrule
\end{tabular}
\caption{Test accuracy of models on different held-out distributions, each consisting of a balanced human / machine split. We observe, somewhat surprisingly, that the OpenAI detector generalizes to Grover generations better than the GLTR detector.}
\label{tab:clf_accuracy}
\end{table}

\subsection{Temporal Analysis}
Using the OpenAI detector scores as a measure of language / page quality, we now characterize the temporal pattern of low quality content on the web. Is there more or less low quality content on the web recently? To control for the fact that the web corpus contains more documents published recently, we plot the ratio between the number of documents detected as low quality and the total number of documents for each historical month.

As visible in Figure \ref{fig:temporal_openai}, we observe a burst in the fraction of low quality documents at the beginning of 2019. One possible explanation for this is in the maturity of technology for fast and realistic-looking generation of text. We will later qualitatively describe these low quality pages.



\subsection{Content Length Analysis}
\begin{figure}
    \centering
    \includegraphics[width=1.0\linewidth]{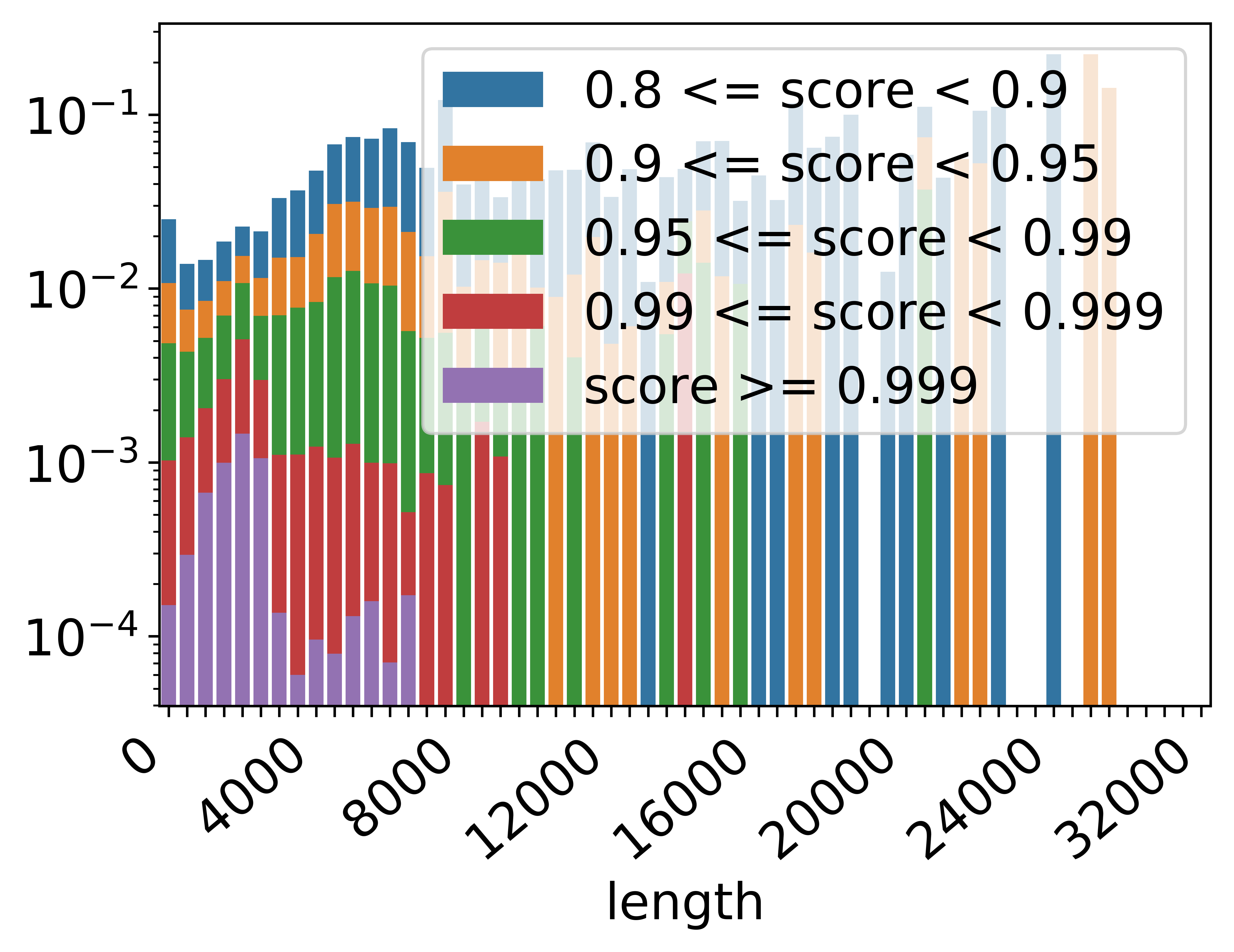}
    \caption{Fraction of documents with OpenAI detector score in different ranges as a function of untruncated character length. Low quality content tends to be shorter in length, peaking at 3000 characters.}
    \label{fig:token_length_openai}
\end{figure}

Figure~\ref{fig:token_length_openai} shows the fraction of documents with OpenAI detector scores in different ranges as a function of \textit{untruncated} document length. Controlling for the fact that shorter documents are more common across the web, we find that low quality content tends to be shorter in length, peaking at about 3000 characters.


\begin{figure*}[]
\centering
    \begin{minipage}[!t]{0.46\linewidth}
    \includegraphics[width=1.0\linewidth]{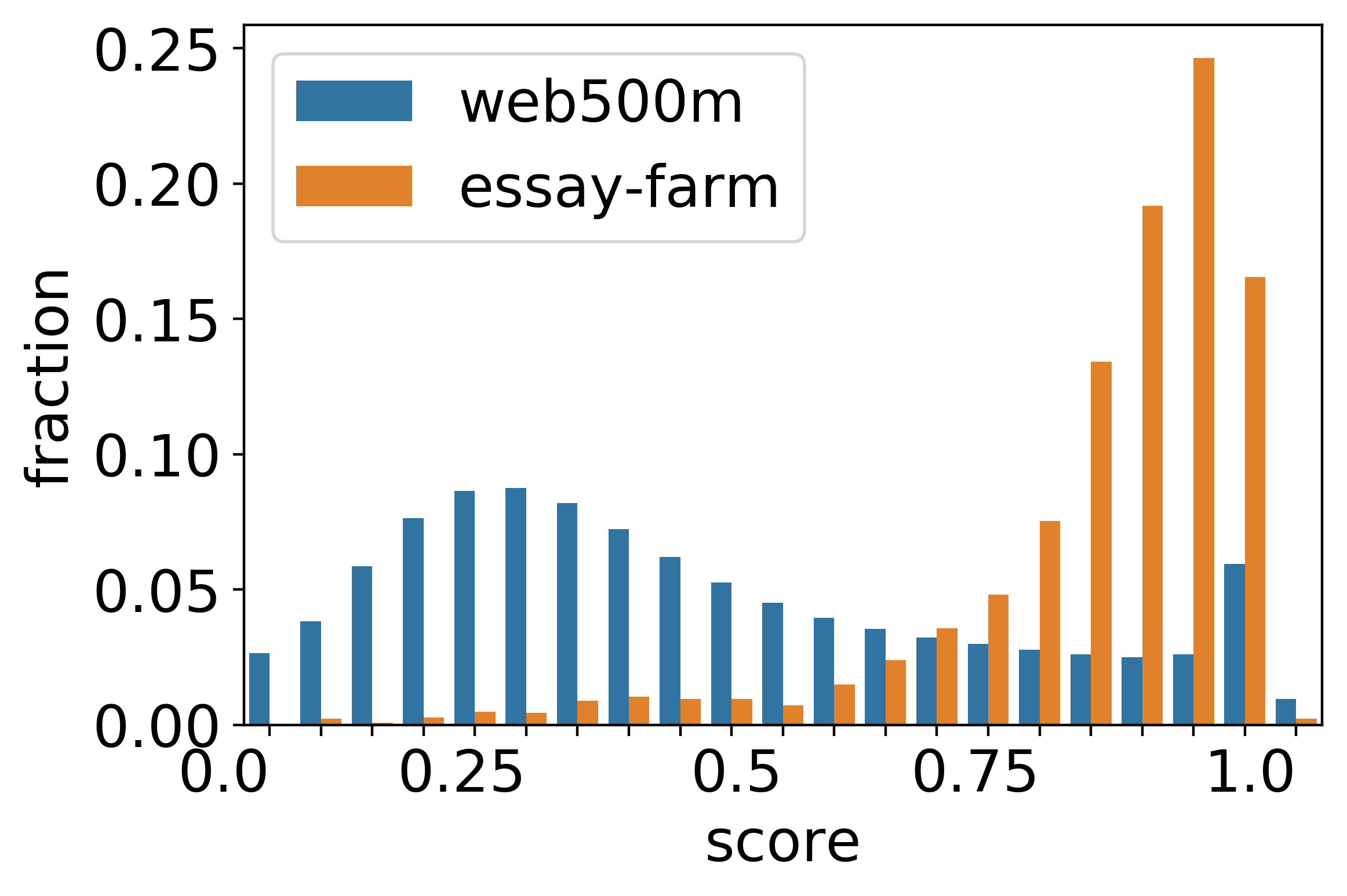}
    \subcaption{GLTR Logistic Regression}
    \end{minipage}%
    \begin{minipage}[!t]{0.46\linewidth}
    \includegraphics[width=1.0\linewidth]{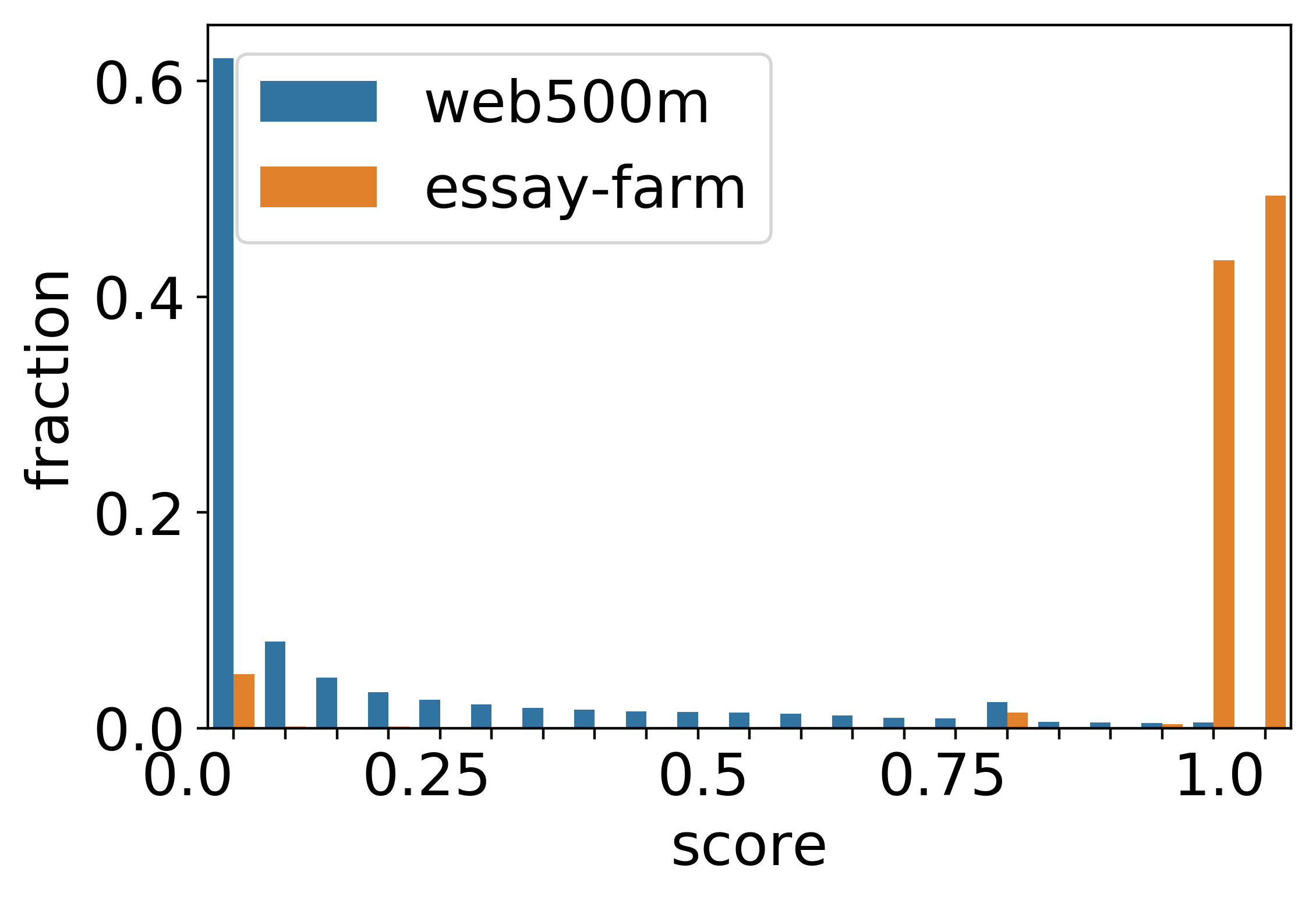}
    \subcaption{OpenAI Detector}
    \end{minipage}%
    \caption{OpenAI detector score distribution on essay writing service domains. The domains have a high incidence of low quality content compared to the rest of the web.}
    \label{fig:essay_farm}
\end{figure*}
\subsection{Topical Analysis}
Figure~\ref{topical_histograms} presents the distribution of OpenAI detector scores by document topic. Topics here are based on the ``Content Categories'' from the Google Cloud Natural Language API \footnote{\url{https://cloud.google.com/natural-language/docs/categories}}. We analyze six high-level topic categories: News, Adult, Law / Government, People / Society, Science and Books / Literature. Based on our empirical results, we find that the score distributions vary significantly based on topic.

Among all topical distributions, we find that a large fraction of documents from the Adult category are flagged as low quality by our detector. Domains such as Games and Books / Literature also tend to be heavier on the low quality side of the histogram. Law / Government and Science topics trend lower with lower quality, suggesting these domains attract higher quality content creators. Food, interesting, is mostly uniform in nature while Health and People / Society follow a convex shape, with large numbers of documents clearly falling near $0$ and $1$.

We found it curious that the Books / Literature domain was flagged as low quality since we would expect it to consist of high quality prose. Upon closer inspection, we found a slew of ``essay farms'', websites trying to sell possibly auto-generated essays to students.
Moreover, the unusual number of documents near $1$ for Health may be likely due to websites selling ``adult health products''.

\subsection{Frequent Terms Analysis}
In this section, we perform an analysis of the frequent terms from documents with different detector scores. In particular, we extract frequent terms for each document in Web500M and present word cloud visualizations for different score ranges.

Figure~\ref{fig:wordcloud} illustrates the most frequent terms for six ranges of detector score: $[0.01, 0.1]$, $[0.1, 0.2]$, $[0.5,0.6]$, $[0.6,0.7]$, $[0.99,1.0]$, and $[0.999,1.0]$.
Based on our observations, the topicality shifts drastically across score ranges. In the low-score range, the top frequent terms are ordinary web applications. However, as the scores approach $1$, we notice heavy occurrences of NSFW terms. To ensure the word clouds do not contain inappropriate language, all clouds for $\mathrm{score} > 0.5$ are computed by first filtering out documents that are marked as NSFW.
Figures~\ref{wordclouds6} and \ref{wordclouds5} are fairly representative of the documents found in these high-score ranges. We make several observations. Firstly, the emergence of the \textit{``essay''}, \textit{``writing''} and \textit{``thesis''} keywords aligns well with the emergence of low quality documents in the Books / Literature topical distribution and is an indicator of the presence of essay farms. Secondly, we find keywords such as \textit{``viagra''} which may explain the low quality peak we observe in the Health topical distribution.
\subsection{Qualitative Analysis}
This section presents key qualitative insights into the kind of web documents our model deems low quality. We manually inspected the top-$0.01\%$ scoring documents. 

\begin{itemize}
    \item \textbf{Machine Translated Text} - We found web documents that look like they might have been translated from English to another language and then back to English.
    \item \textbf{Essay Farms} - We found essay farms selling writing services. Figure \ref{fig:essay_farm} shows the score distribution of pages on a set of essay writing service domains. It's conceivable that some of these pages were machine-generated, although not necessarily by neural generative models.
    \item \textbf{Attempts at Search Engine Optimization (SEO)} - Documents that attempt to perform SEO tend to be flagged as very low quality. This is intuitive since these texts tend to simply string a series of keywords together and are therefore incoherent. Furthermore, we found a moderate number of product pages and professional profiles that also attempt to perform some form of SEO. We observed that media-centric domains, such as image hosting domains, often contain incomprehensible embedded text, possibly for SEO.
    \item \textbf{NSFW (Not-Safe-for-Work) Content} - We observed that a lot of low quality pages contained copious amounts of NSFW content. Upon deeper inspection, we found that many NSFW pages embed long paragraphs of nonsensical text as hidden text. The textual content is generally NSFW and generally incoherent. We speculate that this might also be an attempt at SEO.
\end{itemize}

\section{Conclusion}
This paper posits that detectors trained to discriminate human vs. machine-written text are effective predictors of webpages' language quality, outperforming a baseline supervised spam classifier. We substantiate this through rigorous human evaluation and then apply these low language quality detectors on half a billion webpages. We observed interesting topical and temporal patterns to the low quality content and discovered that many offenders are either (1) machine-translated text, (2) essay farms, (3) attempts at search engine optimization, or (4) NSFW content. We hope researchers interested in text quality find our web-scale analysis useful. Furthermore, we hope they leverage the insight that a reasonable language-quality classifier can be constructed with nothing more than a corpus of human text: train a generative model on the corpus, use it to synthesize machine text, and finally train a model to discriminate between the natural text and synthetic machine text.

\clearpage
\bibliographystyle{ACM-Reference-Format}
\bibliography{wsdm2021}

\end{document}